\documentclass[sigconf]{acmart}

\usepackage{diagbox}  
\usepackage{graphicx} 
\usepackage{xspace}
\usepackage{makecell}
\usepackage{bbding}
\usepackage{booktabs} 
\usepackage{xcolor}
\usepackage{subfig}
\AtBeginDocument{%
  \providecommand\BibTeX{{%
    \normalfont B\kern-0.5em{\scshape i\kern-0.25em b}\kern-0.8em\TeX}}}

\setcopyright{acmcopyright}
\copyrightyear{2020}
\acmYear{2020}
\acmDOI{xxx}

\acmConference[SIGKDD '20]{San Diego '20: SIGKDD}{August 22--27, 2020}{San Diego, CA}
\acmBooktitle{San Diego '20: SIGKDD, August 22--27, 2020, San Diego, CA}
\acmPrice{xxx.00}
\acmISBN{xxx}

\newcommand{\hide}[1]{}

\newcommand{\bit}{\setdefaultleftmargin{2em}{2em}{}{}{}{} \begin{compactitem}}
\newcommand{\eit}{\end{compactitem}}
\newcommand{\ben}{\setdefaultleftmargin{2em}{2em}{}{}{}{} \begin{compactenum}}
\newcommand{\een}{\end{compactenum}}

\newcommand{\eat}[1]{}
\newcommand{\tightlist}{\itemsep=-1pt}

\newcommand{\ie}{{\em i.e.}}
\newcommand{\eg}{{\em e.g.}}

\newcommand{\cal}[1]{\mathcal{#1}}

\newcommand{\scale}{Domain-complexity\xspace}
\newcommand{\noise}{Product-type-variety\xspace}
\newcommand{\sparsity}{Structure-sparsity\xspace}

\newcommand{\doubleBlind}[1]{{\em Omitted, for double-blind rules}}

\newcommand{\graph}{{\textsc{PG}\xspace}}

\newcommand{\method}{{\textsc{AutoKnow}\xspace}}
\newcommand{\methodAbbr}{{\textsc{AK}\xspace}}
\newcommand{\ourTaxo}{\methodAbbr-Taxonomy}
\newcommand{\ourAttr}{\methodAbbr-Relations}
\newcommand{\ourImpute}{\methodAbbr-Imputation}
\newcommand{\ourClean}{\methodAbbr-Cleaning}
\newcommand{\ourSyn}{\methodAbbr-Synonyms}

\newcommand{\compProblem}{\noindent\textbf{Problem definition:}\xspace}

\newcommand{\compTechniques}{\smallskip\noindent\textbf{Key techniques:}\xspace}
\newcommand{\compEvaluation}{\smallskip\noindent\textbf{Component Evaluation:}\xspace}

\setcopyright{rightsretained}

\begin{document}
\fancyhead{}


\title{ \method: Self-Driving Knowledge Collection for \\Products of Thousands of Types}


\author{Xin Luna Dong$^{1}$, Xiang He$^{1}$, Andrey Kan$^{1}$, Xian Li$^{1}$, Yan Liang$^{1}$, Jun Ma$^{1}$, Yifan Ethan Xu$^{1}$, Chenwei Zhang$^{1}$, Tong Zhao$^{1}$, Gabriel Blanco Saldana$^{1}$, Saurabh Deshpande$^{1}$, Alexandre~Michetti~Manduca$^{1}$, Jay Ren$^{1}$, Surender Pal Singh$^{1}$, Fan Xiao$^{1}$, Haw-Shiuan~Chang$^{2\text{*}}$,~Giannis~ Karamanolakis$^{3\text{*}}$,~Yuning~ Mao$^{4\text{*}}$,~Yaqing~Wang$^{5\text{*}}$, Christos~Faloutsos$^{6\text{*}}$,~Andrew~ McCallum$^{2}$,~Jiawei~ Han$^{4}$}\titlenote{Research conducted at Amazon.}
\affiliation{
	\institution{$^1$Amazon  $\quad$ $^2$University of Massachusetts Amherst  $\quad$ $^3$Columbia University}
	\institution{$^4$University of Illinois at Urbana-Champaign  $\quad$ $^5$State University of New York at Buffalo$\quad^6$Carnegie Mellon University}
	\institution{$^1$\{lunadong,xianghe,avkan,xianlee,ynliang,junmaa,xuyifa,cwzhang,zhaoton,
		saldanag,sdeshpa,manduca,renjie,srender,fnxi\}@amazon.com  $\quad$ $^2$\{hschang,mccallum\}@cs.umass.edu  $\quad$ $^3$gkaraman@cs.columbia.edu  $\quad$ $^4$\{yuningm2,hanj\}@illinois.edu  $\quad$ $^5$yaqingwa@buffalo.edu  $\quad$ $^6$christos@cs.cmu.edu  $\quad$ }
}


\renewcommand{\shortauthors}{Dong et al.}

\begin{abstract}
	Can one build a knowledge graph (KG) for all products in the world? Knowledge graphs have firmly established themselves as valuable sources of information for search and question answering, and it is natural to wonder if a KG can contain information about products offered at online retail sites. There have been several successful examples of generic KGs, but organizing information about products poses many additional challenges, including sparsity and noise of structured data for products, complexity of the domain with millions of product types and thousands of attributes, heterogeneity across large number of categories, as well as large and constantly growing number of products.

We describe \method, our automatic (self-driving) system that addresses these challenges.
The system includes a suite of novel techniques for taxonomy construction, product property identification, knowledge extraction, anomaly detection, and synonym discovery. \method\ is
(a) \textbf{automatic}, requiring little human intervention,
(b) \textbf{multi-scalable}, scalable in multiple dimensions (many domains, many products, and
many attributes),
and (c) \textbf{integrative}, exploiting rich customer behavior logs.
\method\ has been operational in collecting product knowledge for over 11K product types.

\end{abstract}

\copyrightyear{2020} 
\acmYear{2020} 
\acmConference[KDD '20]{Proceedings of the 26th ACM SIGKDD Conference on Knowledge Discovery and Data Mining}{August 23--27, 2020}{Virtual Event, CA, USA}
\acmBooktitle{Proceedings of the 26th ACM SIGKDD Conference on Knowledge Discovery and Data Mining (KDD '20), August 23--27, 2020, Virtual Event, CA, USA}
\acmDOI{10.1145/3394486.3403323}
\acmISBN{978-1-4503-7998-4/20/08}





\maketitle

\section{Introduction}
\label{sec:intro}
\begin{figure}[t]
	\centering
	\includegraphics[width=0.49\textwidth]{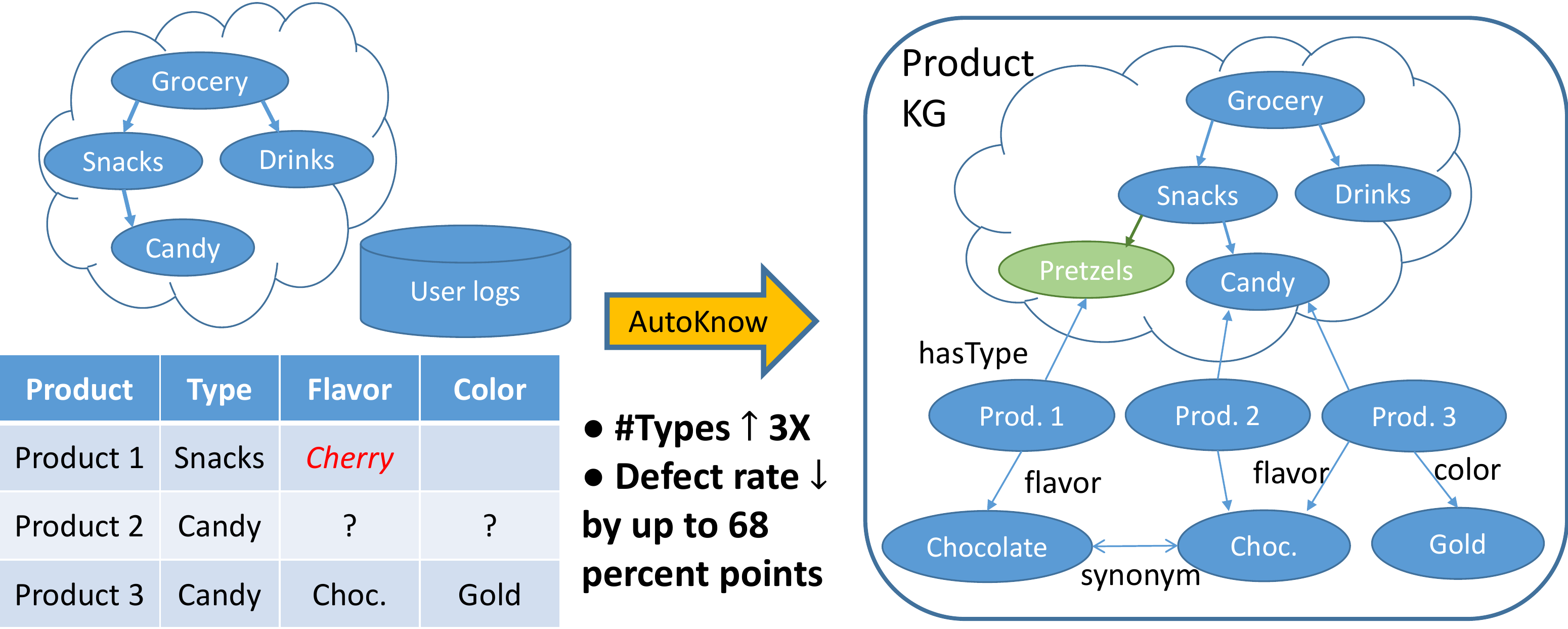}
	\caption{\label{fig:CrownJewel}\small We propose \method, a pipeline that constructs a product knowledge graph. \method\ fixes incorrect values (\eg, ``Flavor:Cherry'' for Product 1) and imputes missing values (\eg, ``Flavor:Choc.'' for Product 2); however, it does not impute where it is inapplicable (\eg, {\em Color} applies for wrapped candies such as Product 3, but does not apply to pretzel snack Product 1). It also extends taxonomy (\eg, {\em Pretzels}) and finds synonyms (\eg, {\em Chocolate} vs. {\em Choc.}).}
        \vspace{-.2in}
\end{figure}

A knowledge graph (KG) describes entities and relations between them; for example, between entities {\em Amazon} and {\em Seattle}, there can be a {\em headquarters\_located\_at} relation. The past decade has witnessed broad applications of KG in search (\eg, by {\em Google} and {\em Bing}) and question answering (\eg, by {\em Amazon Alexa} or {\em Google Home}). {\em How to automatically build a knowledge graph with comprehensive and high-quality data} has been a hot topic for research and industry practice in recent years. In this paper, we answer this question for the {\em Retail Product} domain. Rich product knowledge can significantly improve e-Business shopping experiences through product search, recommendation, and navigation.

Existing industry success for knowledge curation is mainly for popular domains such as {\em Music}, {\em Movie}, and {\em Sport}~\cite{bollacker2008freebase, Gao2018Satori}. Two common features for such domains make them pioneer domains for knowledge collection. First, there are already rich data in structured form and of decent quality for these domains. Taking {\em Movie} as an example, in addition to common knowledge sources such as Wikipedia and WikiData, other authoritative movie data sources include {\em IMDb}~\footnote{https://www.imdb.com/}, and so on. Second, the complexity of the domain schema is manageable. Continuing with the {\em Movie} domain, the Freebase knowledge graph~\cite{bollacker2008freebase} contains 52 entity types and 155 relationships~\cite{zhang2019openki} for this domain. 
An {\em ontology} to describe these types and relationships can be manually defined within weeks, especially by leveraging existing data sources.

The retail product domain presents a set of new challenges for knowledge collection. 

{\bf C1- \sparsity:} First, except for a few categories such as electronics, structured data are sparse and noisy across nearly all data sources. This is because the majority of product data reside in catalogs from e-Business websites such as Amazon, Ebay, and Walmart, and they often rely on data contributed by retailers. In contrast to publishers for digital products like movies and books, in the retail business retailers mainly list product features in titles and descriptions instead of providing structured attribute information, and may even abuse those structured attribute fields for convenience in selling products~\cite{zheng2018opentag,xu2019scaling}. As a result, structured knowledge needs to be mined from textual product profiles (\eg, titles and descriptions). Thousands of product attributes, billions of existing products, and millions of new products emerging on a daily basis, require fully automatic and efficient knowledge discovery and update mechanisms.

{\bf C2- \scale}:
Second, the domain is much more complex. The number of product types is towards millions and there are various relationships between the types like sub-types (\eg, swimsuit vs. athletic swimwear), synonyms (\eg, swimsuit vs. bathsuit), and overlapping types (\eg, fashion swimwear vs. two-piece swimwear). Product attributes vastly differ between types (\eg, compare TVs and dog food), and also evolve over time (\eg, older TVs did not have WiFi connectivity). All of these make it hard to design comprehensive ontology and keep it up-to-date, thus calling for automatic solutions.

{\bf C3- \noise}:
Third, the variety of different product types makes it even harder to train knowledge enrichment and cleaning models. Product attributes, value vocabularies, text patterns in product titles and descriptions often differ for different types. Even neighboring product types can have different attributes; for example, {\em Coffee} and {\em Tea}, which share the same parent {\em Drink}, describe size using different vocabularies and patterns (\eg, ``Ground Coffee, 20 Ounce Bag, Rainforest Alliance Certified'' vs. ``Classic Tea Variety Box, 48 Count (Pack of 1)''). On the one hand, training one single model is inadequate to achieve good results for all different types of products. On the other hand, collecting training data for each of the thousands to millions of product types is extremely expensive, and implausible for less-popular types. Maintaining a huge number of models also brings big system overhead.

With all of these challenges, the solutions in building existing KGs both in industry (\eg, {\em Google Knowledge Graph}, {\em Bing Satori Graph}~\cite{Gao2018Satori}), and in research literature (\eg, Yago~\cite{fabian2007yago}, NELL~\cite{carlson2010toward}, Diadem~\cite{furche2012diadem}, Knowledge Vault~\cite{dong2014knowledge}), cannot directly apply to the retail product domain, as we will further discuss in Section~\ref{sec:related}.

In this paper, we present our solution, which we call \method\ (Figure~\ref{fig:CrownJewel}). {\method} starts with building product type {\em taxonomy} (\ie, types and hypernym relationships) and deciding applicable product attributes for each type; after that, it imputes structured attributes, cleans up noisy values, and identifies synonym expressions for attribute values. 
Imagine how an autonomous-driving vehicle perceives and understands the environment using all the signals available with minimized human interventions. {\method} is {\em self-driving} with the following features. 
\begin{itemize}\tightlist
	\item {\bf Automatic:} First, it trains machine learning models and requires very little manual efforts. In addition, we leverage existing Catalog data and customer behavior logs to generate training data, eliminating the need for manual labeling for the majority of the models and allowing extension to new domains without extra efforts. 
        \item {\bf Multi-scalable:} Our system is scalable in multiple dimensions. It is extensible to new values and is not constrained to existing vocabularies in the training data. It is extensible to new types, as it trains one-size-fits-all models for thousands of types, and the models behave differently for different product types to achieve the best results.
        \item {\bf Integrative:} Finally, the system applies self-guidance, and uses customer behavior logs to identify important product attributes to focus efforts on.
\end{itemize}

A few techniques play a critical role to allow us to scale up to the large number of product types we need to generate knowledge for.
       	First, we leverage the graph structure that naturally applies to knowledge graphs (entities can be considered as nodes and relationships can be considered as edges) and taxonomy (types and hypernym relationships form a tree structure), and apply Graph Neural Network (GNN) for learning. 
       	Second, we take product categorization as input signals to train our models, and combine our tasks with product categorization for multi-task training to allow better performance. 
        Third, we strive to learn with limited labels to alleviate the burden of manual training data creation, relying heavily on weak supervision (e.g., distant supervision) and on semi-supervised learning. 
        Fourth, we mine both facts and heterogeneous expressions for the same concept (\ie, type, attribute value) from customer behavior logs, abundant in the retail domain.

More specifically, we make the following contributions. 

\begin{enumerate}\tightlist
  \item {\bf Operational system:} We describe \method, a comprehensive end-to-end solution for product knowledge collection, covering components from ontology construction and enrichment, to data extraction, cleaning, and normalization. A large part of \method\ has been deployed and operational in collecting over 1B product knowledge facts for over 11K distinct product types, and the knowledge has been used for Amazon search and product detail pages. 
  \item {\bf Technical novelty:} We invented a suite of novel techniques that together allow us to scale up knowledge discovery to thousands of product types. The techniques range from NLP and graph mining to anomaly detection, and leverage state-of-the-art techniques in GNN, transformer, and multi-task learning.
  \item {\bf Empirical study}: We describe our practice on real-world e-Business data from Amazon, showing that we are able to extend the existing ontology by 2.9X, and considerably increase the quality of structured data, on average improving precision by 7.6\% and recall by 16.4\%.
 
\end{enumerate}

Whereas our paper focuses on retail domain and our experiments were conducted on Amazon data, the techniques can be easily applied to other e-Commerce datasets, and adapted to other domains with hierarchical taxonomy, rich text profiles, and customer behavior logs, such as finance, phylogenetics, and biomedical studies.

\section{Definition and System Overview}
\label{sec:overview}

\subsection{Product Knowledge Graph}
A KG is a set of triples in the form of (subject, predicate, object). The {\em subject} is an entity with an ID, and this entity belongs to one or multiple types. The {\em object} can be an entity or an atomic value, such as a string or a number. The {\em predicate} describes the relation between the subject and the object. For example, (prod\_id, hasBrand, brand\_id) is a triple between two entities, whereas (prod\_id, hasSugarsPerServing, ``32'') is a triple between an entity and an atomic value. One can consider the entities and atomic values as nodes in the graph, and predicates as edges that connect the nodes.

For simplicity of problem definition, in this paper we focus on a special type of knowledge graph, which we call a {\em broad graph}. The broad graph is a bipartite graph $G=(N_1, N_2, E)$, where nodes in $N_1$ represent entities of one particular type, called the {\em topic type}, nodes in $N_2$ represent attribute values (that can be entities or atomic values), and edges in $E$ connect each entity with its attribute values. The edges are labeled with corresponding attribute names (Figure~\ref{fig:CrownJewel}). In other words, a broad graph contains only two layers, and thus contains attribute values only for entities of the topic type. We focus on broad graphs where the topic type is product. Once a broad graph is built, one can imagine stacking broad graphs layer by layer to include knowledge about other types of entities (\eg, brand), and eventually arrive at a rich, comprehensive graph.

Product types form a tree-structured {\em taxonomy}, where the root represents all products, each node represents a product type, and each edge represents a sub-type relationship. For example, {\em Coffee} is a sub-type of {\em Drink}.

We assume two sources of input. First, we assume existence of a {\em product Catalog}, including a product taxonomy, a set of product attributes (not necessarily distinguished between different product types), a set of products, and attribute values for each product. We assume that each product has a {\em product profile} that includes title, description, and bullet points, and in addition a set of structured values, where title is required and other fields are optional. Second, we assume existence of {\em customer behavior logs}, such as the query and purchase log, customer reviews, and Q\&A. We next formally define the problem we solve in this paper.

\smallskip
\noindent
    {\bf Problem definition:} Let $\cal{C} = ({\cal T}, {\bf A}, {\bf P})$ be a product Catalog, where (1) $\cal{T}=(\mathbf{T},\mathbf{H})$ denotes a product taxonomy with a set of product types $\bf T$ and the hypernym relationships $\bf H$ between types in $\bf T$; (2) $\bf A$ denotes a set of product attributes, and (3) ${\bf P}=\{\operatorname{PID}, \{T\}, \{(A, V)\}\}$ contains for each product ($\operatorname{PID}$ is the ID) a set of product types $\{T\}$ and a set of attribute-value pairs $\{(A,V)\}$. Let $\cal L$ denote customer behavior logs. 
      {\em Product Knowledge Discovery} takes $\cal C$ and $\cal L$ as input, enriches the product knowledge by adding new types and hypernym relationships to $\cal T$, and new product types and attribute values for each product in $\bf P$.

\begin{figure}[t]
	\centering
	\includegraphics[width=0.45\textwidth]{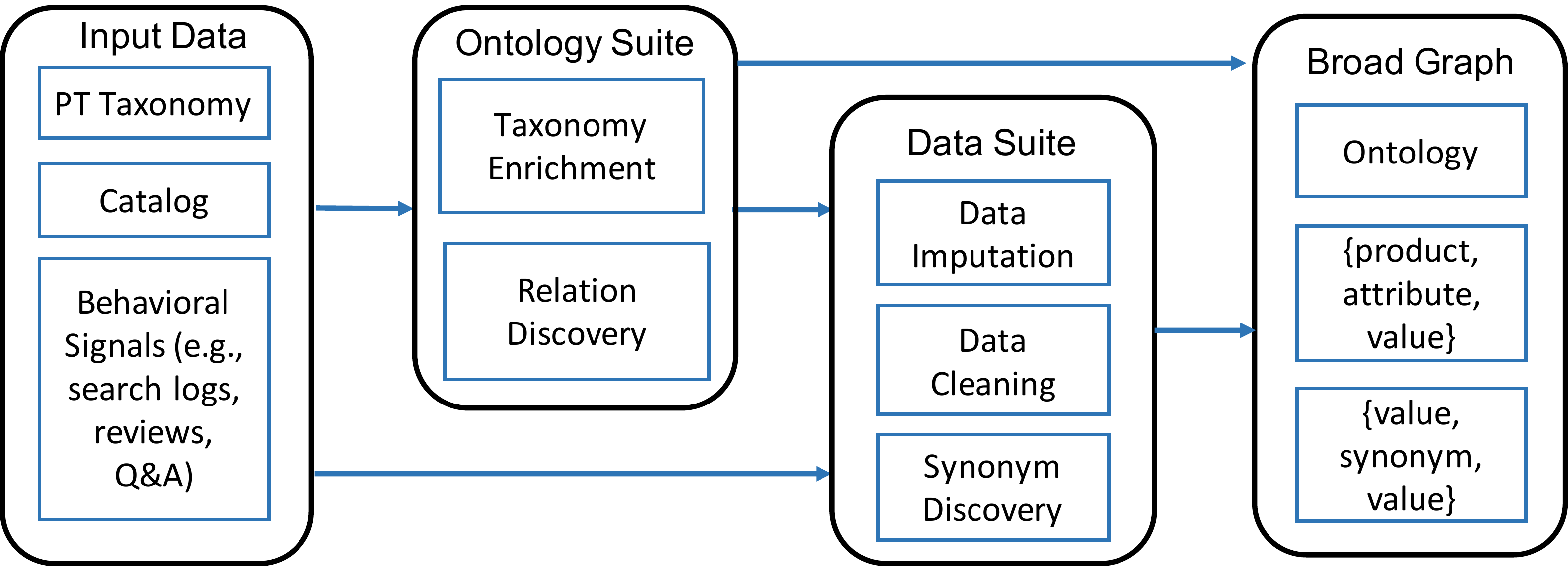}
	\caption{\method\ architecture, containing ontology suite to enrich product ontology and data suite to enrich product structured data.\label{fig:arch}}
        \vspace{-.2in}
\end{figure}

\subsection{System Architecture}
\label{subsec:arch}
Figure~\ref{fig:arch} depicts the architecture of our \method\ system. It has five components, categorized into two function suites. 

{\bf Ontology suite:} The {\em ontology suite} contains two components: {\em Taxonomy enrichment} and {\em Relation discovery}. Taxonomy enrichment identifies new product types not existing in input taxonomy ${\cal T}$ and decides the hypernym relationships between the newly discovered types and existing types, using them to enrich $\cal T$. Relation discovery decides for each product type $T \in \bf T$ and attribute $A \in \bf A$, whether $A$ applies to type $T$ and if so, how important $A$ is when customers make purchase decisions for these products, captured by an {\em importance score}.

{\bf Data suite:} The {\em data suite} contains three components: {\em Data imputation}, {\em Data cleaning}, and {\em Synonym discovery}. Data imputation derives new (attribute, value) pairs for each product in $\bf P$ from product profiles and existing structured attributes. Data cleaning identifies anomalies from existing data in $\bf P$ and newly imputed values. Synonym discovery associates synonyms between product types and attribute values.

Each component is independent, automatic and multi-scalable; on the other hand, the components are well pieced together. The early components provide guidance and richer data for later components; for example, relation discovery identifies important and meaningful relations for the data suite, and data imputation provides richer data for synonym discovery. The later components fix errors from early parts of the pipeline; for example, data cleaning removes mistakes from data imputation.

Table~\ref{tab:system} summarizes how each component of \method\ employs the aforementioned techniques. We next describe in Section~\ref{sec:ontology} how we build the ontology and in Section~\ref{sec:data} how we improve the data. To facilitate understanding of our design choices, for each component we present comparison of our proposed solution with the state-of-the-arts, show ablation study, and show real examples in Appendix~\ref{sec:appendix-example}. Unless otherwise mentioned, we use the {\em Grocery} domain in the US market and the {\em flavor} attribute to illustrate our results, but we have observed the same trend throughout different domains and attributes. We describe detail of the experimental setup and end-to-end empirical study in Section~\ref{sec:exp} and lessons we learned in Section~\ref{sec:disc}.
\begin{table}[t]
	\begin{center}
		\caption{Key techniques employed by each component.
			\label{tab:system}}
                \vspace{-.1in}
		\begin{tabular}{ l| c | c||c| c | c|}
			\diagbox{Techniques}{Component}    & \rotatebox{80}{\ourTaxo} & \rotatebox{80}{\ourAttr} & \rotatebox{80}{\ourImpute} & \rotatebox{80}{\ourClean} & \rotatebox{80}{\ourSyn} \\
			\hline
			Graph structure &  X  &  & X & &  \\
			Taxonomy signal &  & & X & X &  \\
			Distant supervision & X  &  & X & X &  \\
			Behavior information & X  & X  & & & X \\ \hline
		\end{tabular}
                \vspace{-.1in}
	\end{center}
\end{table}

\section{Enriching the Ontology}
\label{sec:ontology}


\subsection{Taxonomy Enrichment}
\label{sec:taxonomy}

\compProblem
Online catalog taxonomies are often built and maintained manually by human experts (\eg, taxonomists), which is labor-intensive and hard to scale up, leaving out fine-grained product types. {\em How to discover new product types and attach them to the existing taxonomy in a scalable, automated fashion} is critical to address the \textbf{C2- \scale} challenge.

Formally, given an existing product taxonomy ${\cal T}=({\bf T}, {\bf H})$, {\em Taxonomy Enrichment} extends it with ${\cal T}'=({\bf T} \cup {\bf T'}, {\bf H} \cup {\bf H'})$, where ${\bf T'}$ is the set of new product types, and ${\bf H'}$ is the additional set of hypernym relationships between types in ${\bf T}$ and in ${\bf T'}$.

\compTechniques
The product domain proposes its unique challenges for taxonomy construction. A product type and its instances, or a type and its sub-types, are unlikely to be mentioned in the same sentence as in other domains such as ``big US cities like Seattle'', so traditional methods like Hearst patterns do not apply. Our key intuition is that since product types are very important, they are frequently mentioned in product titles (see Table~\ref{tab:tagging-example} for an example) and search queries (\ie, ``{\em k-cups} dunkin donuts dark''); we thus leverage existing resources such as product profiles in the Catalog $\cal C$ or search queries in behavior logs $\mathcal{L}$ to effectively supervise the taxonomy enrichment process. 

We enrich product taxonomy in two steps. 
We first discover new types $\bf{T'}$ from product titles or customer search queries by training a type extractor. 
Then, we attach candidate types in $\bf{T'}$ to the existing taxonomy ${\cal T}$ by solving a hypernym classification problem.
We next briefly describe high-level ideas of each step and details can be found in~\cite{mao2020octet}. 
\begin{table}[t]
    \centering
    \caption{Example of input(text)/output(BIOE tag) sequences for the type and flavor of an ice cream product.}\label{tab:tagging-example}
    \vspace{-.1in}
    \resizebox{\columnwidth}{!}{
    \begin{tabular}{|c|c|c|c|c|c|c|c|c|}
    \hline
       \textbf{Input}  & Ben & \& & Jerry's & black & cherry & cheesecake & ice & cream \\\hline
      \textbf{Output}  & O & O & O & B-flavor & I-flavor & E-flavor & B-type & E-type \\\hline
    \end{tabular}}
    \vspace{-.2in}
\end{table}

\smallskip
\noindent
\underline{Type extraction:} Type extraction discovers new product types mentioned in product titles or search queries.
For the purpose of recognizing new product types from product titles, it is critical that we are able to extract types not included in training data.
Thus, we adopt an open-world tagging model and formulate type extraction as a ``BIOE'' sequential labeling problem.
In particular, given the product's title sequence $(x_1, x_2, ..., x_L)$, the model outputs the sequence of $(y_1, y_2, ..., y_L)$, where $y_i\in\{B, I , O, E\}$, representing "begin", "inside", "outside", "end" respectively. Table~\ref{tab:tagging-example} illustrates an example of sequential labels obtained using OpenTag \cite{zheng2018opentag}: ``ice cream'' is labeled as product type, and "black cherry cheesecake" as product flavor. 


To train the model, we adopt distant supervision to generate the training labels. For product titles, we look for product types in Catalog provided by retailers (restricted to the existing product types), and generate BIOE tags when types are explicitly and exactly mentioned in their titles.
For queries, we look for the type of purchased products in the query to generate BIOE tags.
Once the extraction models are trained, we apply them on product titles and queries. New types from titles are taken as ${\bf T'}$, and those from queries, albeit noisier, will be used for type attachment. 

\smallskip
\noindent
\underline{Type Attachment:} Type attachment organizes extracted types into the existing taxonomy. We thus solve a binary classification problem, where the classifier
determines if the hypernym relationship exists between two types $T\in{\bf T}, T'\in{\bf T}'$.

Our key intuition is to capture various signals from customer behaviors with a GNN-based module. 
In particular, we first construct a graph where the nodes represent types, products, and queries, and the edges represent various relationships including 1) product co-viewing, 2) a query leading to a product purchase, 3) the type mentioned in a query or a product (according to the extraction). The GNN model allows us to refine the node representation using the neighborhood information on the graph. Finally, the type representation for each $T\in {\bf T}\cup{\bf T'}$ is combined with semantic features (\eg, word embedding) of the type names and fed to the classifier. 

To train the model, we again apply distant supervision. We use the type hypernym pairs in the existing taxonomy as the supervision to generate positive labels, and generate five negative labels by randomly replacing the hyponym type with other product types. 

\compEvaluation For product type extraction in the Grocery domain, we obtained 87.5\% precision according to MTurk evaluation; in comparison to state-of-the-art techniques, Noun Phrase (NP) chunking obtains 12.3\% precision and AutoPhrase~\cite{shang2018automated} obtains 20.9\% precision.


For type attachment, we took hypernym relationships from existing taxonomy as ground truth, randomly sampled 80\% for model training, 10\% as validation set and 10\% for testing. 
We measured both Edge-F1 (F-measure on parent-child relationship) and Ancestor-F1 (F-measure on ancestor-child relationship)~\cite{bansal2014structured,mao2018end}. Table~\ref{tab:taxo-baseline} shows that our GNN-based model significantly outperforms the state-of-the-art baselines, improving Edge-F1 by 54.3\% over HiDir~\cite{wang2014hierarchical}, and by 17.7\% over MSejrKu~\cite{schlichtkrull2016msejrku}. Ablation tests show that both the multi-hop GNN and the user behavior increase the performance.
\begin{table}[t]
    \caption{{\ourTaxo} improves over state-of-the-art by 17.7\% on Edge-F1.}
    \label{tab:taxo-baseline}
    \centering
    \vspace*{-.3cm}
    \scalebox{.88}{
    \begin{tabular}{|l|c|c|}
        \hline
         \textbf{Method} & \textbf{Edge-F1} & \textbf{Ancestor-F1}\\
        \hline
        Substr~\cite{bordea2016semeval} & 10.7 & 52.9 \\
        HiDir~\cite{wang2014hierarchical} & 40.5 & 66.4 \\
        MSejrKu~\cite{schlichtkrull2016msejrku}  & 53.1 & 76.7 \\\hline
        Type-Attachment & \textbf{62.5} & \textbf{84.2} \\
        ~~ w/o. multi-hop ($\ge$2) GNN & 50.4 ($\downarrow$12.1\%) & 75.9 ($\downarrow$8.3\%)\\
        ~~ w/o. user behavior (query$\leftrightarrow$product) & 60.1 ($\downarrow$2.4\%) & 83.0 ($\downarrow$1.2\%)\\

        \hline
    \end{tabular}
    }
\vspace*{-.25in}
\end{table}

\subsection{Relation Discovery}
\label{sec:attributes}
\compProblem
In a catalog there are often thousands of product attributes; however, different sets of attributes apply to different product types (\eg, {\em flavor} applies to snacks, but not to shampoos), and among them, only a small portion have a big influence on customer shopping decisions (\eg, {\em brand} is more likely to affect shopping decisions for snacks, but less for fruits). Understanding applicability and importance will help filter values for non-applicable attributes and prioritize enrichment and cleaning for important attributes. Thus, {\em how to identify applicable and important attributes for thousands of types} is another key problem to solve to address the \textbf{C2- \scale} challenge.



Formally, given a product taxonomy ${\cal T}=({\bf T}, {\bf H})$ and a set of product attributes $\bf A$, {\em Relation Discovery} decides for each $(T, A)\in{\bf T}\times{\bf A}$, (1) whether attribute $A$ applies to products of type $T$, denoted by $(T, A)\rightarrow\{0, 1\}$, and (2) how important $A$ is for purchase decisions on products of $T$, denoted by $(T, A)\rightarrow[0,1]$. Here, we do not consider newly extracted types in $\bf T'$, since they are often sparse.




\compTechniques
Intuitively, important attributes will be frequently mentioned by sellers and buyers, whereas inapplicable attributes will appear rarely. Previous approaches explored this intuition, but either leveraged only one text source at a time (\eg, only customer reviews) or combined sources according to a pre-defined rule~\cite{hopkinson2018demand,sun2018important,razniewski2017doctoral}. Here we train a classification model to decide attribute applicability, and a regression model to decide attribute importance. We used Random Forest for both models and employ two types of features reflecting behavior of the customers.
\begin{itemize}
  \item {\em Seller behavior}, captured by coverage of attribute values for a particular product type, and frequency of mentioning attribute values in product profiles.
  \item {\em Buyer behavior}, captured by frequency of mentioning attribute values in search queries, reviews, Q\&A sessions, etc.
\end{itemize}


For a given $(T, A)$ pair, we compute features that correspond to different signals (\eg, reviews, search logs, etc.). To this end, we estimate frequencies of mentions of attribute values in the corresponding text sources (see details in Appendix~\ref{sec:appendix-attribute}).
Note that sellers are required to provide certain applicable attributes (\eg, {\em barcode}). These attributes have high coverage, but they are not always important for shopping decisions and appear rarely in customer logs. We thus train two different models for applicability and importance to capture such subtle differences.

We collect manual annotations for training, both in-house and using MTurk. In the latter case, for a given $(T, A)$ pair, we asked six MTurk workers whether the attribute $A$ applies to products of type $T$, and how likely $A$ will influence shopping decisions for products of type $T$. The applicability is decided by majority voting, and importance is decided by averaging influence likelihood grades. Once we trained the model, we apply it to all $(T, A)$ pairs to decide applicability and importance.

\compEvaluation
We collected two datasets. The first dataset contains 807 applicability and importance labels for 11 common attributes (\eg, {\em brand, flavor, scent}) and 79 randomly sampled product types. The second dataset contains 240 applicability labels for 7 product types (\eg, {\em Shampoo, Coffee, Battery}) and 180 attributes for which there are values in the Catalog. We combined the data, used 80\% for training and 20\% testing, and reported results in Table~\ref{tab-applicable}. Our results show that comparing with the strongest signal--coverage, various behavior signals improved F-measure by 4.7\% for applicability, and improved Spearman for importance by 1.9X. Ablation tests show that both buyer features and seller features contribute to the final results.


\begin{table}[t]
	\centering
	\caption{\ourAttr{} outperforms using only coverage features on both applicability prediction (by 4.7\%) and importance prediction (1.9X). Here ``Seller features'' does not include the ``Coverage features''. \label{tab-applicable}}
        \vspace{-.1in}
		\begin{tabular}{|l|c|c|c|c|}
			\hline
			{\bf Method} & {\bf Precision} & {\bf Recall} & {\bf F1} & {\bf Spearman}\\
			\hline
			Coverage features & 0.90 & 0.80 & 0.85 & 0.39 \\
			Seller features  & 0.90 & {\bf 0.84} & 0.87 & 0.72 \\
			Buyer features   & 0.86 & 0.83 & 0.84 & 0.68 \\
			All features     & {\bf 0.94} & {\bf 0.84} & {\bf 0.89} & {\bf 0.74} \\
			\hline
	\end{tabular}
        \vspace{-.25in}
\end{table}

\hide{
\begin{table}[t]
	\centering
	\caption{\ourAttr{} outperforms using single coverage feature on both applicability prediction and importance prediction (2.7X). Here ``seller features'' does not include the coverage feature. \label{tab-datasets}}
        \vspace{-.2in}
	\resizebox{\columnwidth}{!}{
		\begin{tabular}{|c|c|c|c|c|}
			\hline
			& \multicolumn{4}{|c|}{{\em Attr-focused}: 11 attributes, 79 PTs, 807 labels } \\
			\hline
			{\bf method} & {\bf precision} & {\bf recall} & {\bf F1} & {\bf Spearman}\\
			\hline
			coverage feature & 0.85 & 0.88 & 0.87 & 0.33 \\
			seller features  & 0.85 & 0.87 & 0.86 & 0.72 \\
			buyer features   & 0.85 & 0.86 & 0.85 & 0.69 \\
			all features     & {\bf 0.88} & {\bf 0.90} & {\bf 0.89} & {\bf 0.74} \\
			\hline
			& \multicolumn{4}{|c|}{{\em Type-focused}: 180 attributes, 7 PTs, 240 labels } \\
			\hline
			{\bf method} & {\bf precision} & {\bf recall} & {\bf F1} & {\bf Spearman} \\
			\hline
			coverage feature & {\bf 0.90} & 0.59 & 0.72 & n/a\\
			seller features  & 0.87 & 0.62 & 0.73 & \\
			buyer features   & 0.88 & {\bf 0.72} & {\bf 0.79} & \\
			all features     & 0.87 & 0.62 & 0.73 & \\
			\hline
	\end{tabular}}
	\vspace{-.3in}
\end{table}
}


\section{Enriching and Cleaning Knowledge}
\label{sec:data}
\subsection{Data Imputation}
\label{sec:imputation}
\compProblem
The Data Imputation component addresses the \textbf{C1- \sparsity} challenge by extracting structured values from product profiles to increase coverage.
Formally, given product information $(\operatorname{PID}, \{T\}, \{(A, V)\})$, {\em Data imputation} extracts new $(A, V)$ pairs for each product from its profiles (\ie, title, description, and bullets). 

State-of-the-art techniques have solved the problem for a type-attribute pair $(T, A)$, obtaining high extraction quality with BIOE sequential labeling combined with active learning~\cite{zheng2018opentag}. Equation~(\ref{eqn:bioe}) shows  sequential labeling with BiLSTM and CRF:
\vspace{-.05in}
\begin{equation}
\label{eqn:bioe}
(y_1,y_2,...y_L)=\operatorname{CRF}(\operatorname{BiLSTM}(e_{x_1}, e_{x_2}, ..., e_{x_L})),
\end{equation}
where $e_{x_i}$ is the embedding of $x_i$, usually initialized with pre-trained word embedding such as GloVe~\cite{pennington2014glove}, and fine-tuned during model training. As an example, the output sequence tags in Table \ref{tab:tagging-example} shows that "black cherry cheesecake'' is a flavor of the product.

However, the technique does not scale up to thousands to millions of product types and tens to hundreds of attributes that apply to each type. {\em How to train an extraction model that acknowledges the differences between different product types} is critical to scale up sequential labeling to address the {\bf C3- \noise} challenge.

\compTechniques
We proposed an innovative taxonomy-aware sequence tagging approach that makes predictions conditioned on the product type. We describe the high-level ideas next and details can be found in~\cite{karamanolakis2020txtract}.

We extended sequence tagging described in Equation~(\ref{eqn:bioe}) in two ways. First, we condition model predictions on product type $T\in\bf T$:
\begin{equation}
(y_1,y_2,...y_L)=\operatorname{CRF}(\operatorname{CondSelfAtt}(\operatorname{BiLSTM}(e_{x_1}, e_{x_2}, ..., e_{x_L}), e_T))
\end{equation}
where $e_T$ is the pre-trained hyperbolic-space embedding (Poincare~\cite{nickel2017poincare}) of product type $T$, known to preserve the hierarchical relation between taxonomy nodes. $\operatorname{CondSelfAtt}$ is the conditional self attention layer that allows $e_T$ to influence the attention weights.

Second, to better identify tokens that indicate the product type, and address the problem that products can be mis-classified or product type information can be missing in a catalog, we employ multi-task learning: training sequence tagging and product categorization at the same time, with a shared $\operatorname{BiLSTM}$ layer.

We again adopt the distant supervision approach to automatically generate the training sequence labels by text matching between product profiles and available attribute values in the Catalog. The trained model is then applied to all $(\operatorname{PID}, A)$ pairs for predicting missing values. 

\compEvaluation
In Table~\ref{tab:flavor-results-detailed}, we show the performance evaluation and ablation studies of \textit{flavor} extraction across 4000 types of products in the Grocery domain. Compared to the baseline BiLSTM-CRF model adopted by current state-of-the-art~\cite{zheng2018opentag}, both CondSelfAtt and MultiTask learning, when applied alone, improve F1 score by at least 7.0\%; combination of the two together improved F1 by 10.1\%.

\begin{table}[t]
\caption{{\ourImpute} improves over state-of-the-art by 10.1\% on F1 for \emph{flavor} extraction across 4,000 types.
  \label{tab:flavor-results-detailed}}
\vspace{-.1in}
\begin{tabular}{|l|cccc|}
\hline
\textbf{Model} & \textbf{Vocab size} & \textbf{Precision}      & \textbf{Recall} & \textbf{F1} \\
\hline
BiLSTM-CRF~\cite{zheng2018opentag} & 6756 & 70.3 & 49.6              &  57.5 \\
\hline
\ourImpute &\textbf{13093}  &  70.9 & \textbf{57.8} & \textbf{63.3}  \\ 
~~~~ w/o. CondSelfAtt &9528 & \textbf{74.5} & 53.2  &61.5 \\
~~~~ w/o. MultiTask & 12558 & 68.8 & 57.0 & 61.9  \\
\hline 
\end{tabular}
\vspace{-.2in}
\end{table}

\subsection{Data Cleaning}
\label{sec:cleaning}
\compProblem
The structured data contributed by retailer are often error-prone because of misunderstanding or intentional abuse of the product attributes. Detecting anomalies and removing them is thus another important aspect to address the \textbf{C1- \sparsity} challenge.
Formally, given product information $(\operatorname{PID}, \{T\}, \\\{(A,V)\})$, {\em Data cleaning} identifies $(A, V)$ pairs that are incorrect for the product, such as $(A=\text{flavor}, V=\text{\em ``1 lb. box''})$ for a box of chocolate and $(A=\text{color}, V=\text{\em ``100\% Cotton''})$ for a shirt.

Abedjan et al.~\cite{abedjan2016detecting} have made successes in cleaning values of types like numerical and date/time. We focus our discussion on textual attributes, which are often considered as most challenging in cleaning. Similar to data imputation, the key is to address the {\bf C3- \noise} challenge such that we can scale up to nearly millions of types. In particular, we need to answer the question: {\em how to identify anomaly values inconsistent with product profiles for a large number of product types?}

\compTechniques
Our intuition is that an attribute value shall be consistent with the contexts provided by the product profiles. 
We propose a transformer-based \cite{vaswani2017attention} neural net model that jointly processes signals from textual product profile ($D$) 
and the product taxonomy $T$ via a multi-head attention mechanism to decide if a triple $(\operatorname{PID}, A, V)$ is correct (\ie, whether $V$ is the correct value of attribute $A$ for product $\operatorname{PID}$). 
The model is capable of learning from raw textual input without extensive feature engineering, making it ideal for scaling to thousands of types.

The raw input of the model is the concatenation of token sequences in $D$, $T$ and $V$. For the $i$-th token in the sequence, a learnable embedding vector $\boldsymbol{e}_i$ is constructed by summing up three embedding vectors of the same dimension: 
\begin{equation}
\label{eq:cleaning}
\boldsymbol{e}_i = \boldsymbol{e}^{\text{FastText}}_i + \boldsymbol{e}^{\text{Segment}}_i + \boldsymbol{e}^{\text{Position}}_i,
\end{equation}
where $\boldsymbol{e}^{\text{FastText}}_i$ is the pre-trained FastText embedding \cite{bojanowski2017enriching} of token $i$, $\boldsymbol{e}^{\text{Segment}}_i$ is a segment embedding vector that indicates to which source sequence ($D$, $T$ or $V$) token $i$ belongs, and $\boldsymbol{e}^{\text{Position}}_i$ is a positional embedding \cite{vaswani2017attention} that indicates the relative location of token $i$ in the sequence. The sequence of embeddings $[\boldsymbol{e}_1, \boldsymbol{e}_2, \ldots]$
is propagated through a multi-layer transformer model whose output embedding vector $\boldsymbol{e}^{Out}$ captures the distilled representations of all three input sequences. Finally, $\boldsymbol{e}^{Out}$ passes through a dense layer followed by a sigmoid node to produce a single score between $0$ and $1$, indicating the consistency of $D$ and $V$; in other words, the likelihood of the input triple $(\operatorname{PID}, A, V)$ being correct (see Appendix \ref{appendix:cleaning} for details and Figure \ref{fig:cleaning} for the model architecture).

To train the cleaning model, we adopt distant supervision to automatically generate training labels from the input Catalog.  We generate positive examples by selecting high-frequency values that appear in multiple brands, then for each positive example we randomly apply one of the following three procedures to generate a negative example: 1) We build a vocabulary $\operatorname{vocab}(A)$ for each attribute $A$ and replace a catalog value $V$ of $A$ with a randomly selected value from $\operatorname{vocab}(A)$; 2) We randomly select $n$-grams from the product title that does not contain the catalog value $V$, where $n$ is a random number drawn according to the distribution of lengths of tokens in $\operatorname{vocab}(A)$; 3) We randomly pick the value of another attribute $A' \ne A$ to replace $V$. At inference time, we apply our model to every $(\operatorname{PID}, A, V)$ triple and consider those with a low confidence as incorrect.

\compEvaluation
As shown in Table \ref{tab:cleaning}, evaluation on the {\em flavor} attribute for the {\em Grocery} domain on 2230 labels across 223 types shows that our model improves PRAUC over state-of-the-art anomaly detection technique~\cite{liu2008isolation} by 75.3\%, and considering the product taxonomy in addition to product profiles improved PRAUC by 6.7\%. 

\begin{table}[t]
	\centering
	\caption{\ourClean\ improves over state-of-the-art anomaly detection by 75.3\% on PRAUC. R@.7P shows the recall when the precision is 0.7, etc. \label{tab:cleaning}}
        \vspace{-.1in}
	\resizebox{\columnwidth}{!}{
	\begin{tabular}{|l|ccccc|}
		\hline
		{\bf Model} &  {\bf PRAUC} & {\bf R@.7P } &  {\bf R@.8P} &  {\bf R@.9P } & {\bf R@.95P} \\
		\hline
		Anomaly Detection~\cite{liu2008isolation} & 32.0 & 2.4 & 1.3 & 1.3 & 1.3 \\
                \hline
		\ourClean & \textbf{56.1} & \textbf{59.6} & \textbf{39.8} & \textbf{26.0} & \textbf{20.7} \\
		~~~~w/o. Taxonomy & 52.6 & 52.6 & 36.2 & 22.4 & 3.0\\\hline
	\end{tabular}}
        \vspace{-.3in}
\end{table}

\subsection{Synonym Finding}
\label{sec:synonym}
We finally very briefly discuss how we identify synonyms with the same semantic meaning, including spelling variants (\eg, {\em Reese's} vs. {\em reese}), acronyms or abbreviation (\eg, {\em FTO} vs. {\em fair trade organic}), and semantic equivalence (\eg, {\em lite} vs. {\em low sugar}). Aligning synonym values is another important aspect to address the \textbf{C1- \sparsity} challenge, and {\em how to train a domain-specific model to distinguish identical values and highly-similar values} is a key question to answer.

Our method has two stages. First, we apply collaborative filtering~\cite{linden2003amazon} on customer co-view behavior signals to retain product pairs with high similarity, and take their attribute values as candidate pairs for synonyms. Such a candidate set is very noisy, hence requires heavy filtering. Second, we train a simple logistic regression model to decide if a candidate pair has exactly the same meaning. The features we consider include edit distance, pre-trained MT-DNN model~\cite{liu2019mt-dnn} score, and features regarding {\em distinct vs. common words}. The features regarding distinct vs. common words play a critical role in the model; they focus on three sets of words:  words appearing only in the first candidate but not the second, and vice versa, and words shared by the two candidates. Between every two out of these three sets, edit distance and embedding similarity are computed and used as features.

An experiment on 2500 candidate pairs (143 positive; half used for training) shows a PRAUC of 0.83 on Grocery flavor; removing the distinct-word features will reduce the PRAUC to 0.79.

\section{Experimental Results}
\label{sec:exp}
We now present our product knowledge graph (\graph{}) produced by the \method{} pipeline. We show that we built a graph with over 1 billion triples for over 11K product types and significantly improved accuracy and completeness of the data. Note that we have already compared each individual component with state-of-the-art in previous sections, so here we only compare \graph\ with the raw input data. 

\subsection{Input Data and Resulting Graph}
\begin{table}[t]
  \caption{Statistics for raw data used as input to \method.\label{tab:cat-stats} }
        \vspace{-.1in}
	\begin{tabular}{|c|c|c|c|c|}
		\hline
		{\bf Product Domain} & {\bf Grocery} & {\bf Health} & {\bf Beauty} & {\bf Baby} \\
		\hline
		\#types & 3,169 & 1,850 & 990 & 697 \\
		med. \# products/type & 1,760 & 18,320 & 27,150 & 28,700 \\
		\#attributes  & 1,243 & 1,824 & 1,657 & 1,511 \\
		med. \#attrs/type   &   113 &   195 & 228 & 206 \\
		\hline
	\end{tabular}
        \vspace{-.2in}
\end{table}

\eat{
\begin{figure}[t]
 	\centering
 	\includegraphics[width=0.6\linewidth]{FIG/num-products-vs-num-types.pdf}
 	\caption{Distribution of products by product types is highly skewed. Most product types have fewer than $1,000$ products.}
 	\label{fig:type-sizes}
\end{figure}
}

\noindent\textbf{Raw Data:} \method{} takes Amazon Product Catalog, including the product type taxonomy, as input. We chose products of four domains (\ie, high-level categories): Grocery, Health, Beauty, and Baby. These domains have the commonality that they contain quite sparse structured data; on the other hand, the numbers of types and the sizes vary from domain to domain. We consider products that have at least on page view in a randomly chosen month. 

Table~\ref{tab:cat-stats} shows statistics of the domains. For each domain, there are hundreds to thousands of types in the Catalog taxonomy, and the median number of products per type is thousands to tens of thousands. 
Amazon Catalog contains thousands of attributes; however, for each individual product most of the attributes do not apply. Thus for each product, there are typically tens to hundreds of populated values. We say an attribute is covered by a type if at least one product of that type has a value in Catalog for the attribute. As shown in the statistics, each domain involves thousands of attributes, and the median number of attributes per product type is 100-250.

\smallskip
\noindent\textbf{Building a Graph:}\xspace
We implemented \method{} in a distributed setting using Apache Spark 2.4 on an Amazon EMR cluster, and Python 3.6 on individual hosts. Deep learning was implemented using TensorFlow and AWS Deep Graph Library~\footnote{https://www.dgl.ai/} was used to implement Graph Neural Networks for \ourTaxo{}. \ourAttr{} component was implemented using Spark ML. \ourImpute{} component used an AWS SageMaker instance for training\footnote{https://aws.amazon.com/sagemaker/}. 

\smallskip
\noindent\textbf{Resulting \graph:}\xspace
Key characteristics of our \graph{} are shown in Table~\ref{tab:pg-stats}. The table presents aggregated statistics for the four product domains. We highlight that after product type extraction, we increase the size of the taxonomy by 2.9X, from 6.7K to 19K; some types appear in different domains and there are 11K unique types. We show how much we improve the quality of the structured knowledge in the next section.

\begin{table}
	\caption{Aggregated statistics describing our \graph{} on four product domains (Grocery, Baby, Beauty, Health).\label{tab:pg-stats} }
        \vspace{-.1in}
	\begin{tabular}{|c|c|c|c|}
		\hline
		{\bf \#Triples} & {\bf \#Attributes} & {\bf \#Types} & {\bf \#Products} \\
		\hline
		>1B & >1K & >19K & >30M \\
		\hline
	\end{tabular}
\end{table}

\subsection{Quality Evaluation}
\noindent\textbf{Metrics:}
We report {\em precision}, {\em recall}, {\em F-metric} of the knowledge. To understand how much gap there is in providing correct structured data for each attribute, we also define a new metric, called {\em defect rate}, the percentage of (product, attribute) pairs with missing or incorrect values.
Specifically, consider an attribute $A$. Let $c$ be the number of products with correct values for $A$, $w$ be the number of products with a wrong value for $A$, $s$ be the number of products where $A$ does not apply but there is a value (\eg, flavor for shoes), $m$ be the number of products where $A$ applies but the value is missing, and $t$ be the number of products within the scope. We compute {\em applicability}, the percentage of products where $A$ applies, as $(c+w+m)/t$; {\em coverage} as $(c+s+w)/t$; {\em precision} as $c/(c+s+w)$; {\em recall} as $c/(c+w+m)$; and {\em defect rate} as $D=(w+s+m)/(c+w+s+m)$.

We consider three types of triples: triples with product types such as (product-1, hasType, Television), triples with attribute values such as (product-2, hasBrand, Sony), and triples depicting entity relations such as (chocolate, isSynonymOf, choc). We report precision for each type of triples. Computing recall is hard, especially for type triples and relation triples, since it is nearly impossible to find all applicable types and synonyms; we thus only report it for triples with attribute values. For triples involving products, we used product popularity weighted sampling.

\begin{table}
	\caption{\method\ achieved 87.7\% type precision and increased the number of types by 2.9X. \label{tab:res-types} }
        \vspace{-.1in}
	\begin{tabular}{|l|c|c|c|c|c|}
		\hline
		& {\bf Grocery} & {\bf Health} & {\bf Beauty} & {\bf Baby} & {\bf Avg}\\
		\hline
		precision & 93.89\%	& 84.60\% & 82.24\% & 89.97\% & 87.68\% \\
        MoE & 2.71\% & 4.08\% & 4.32\% & 3.40\% & 3.63\% \\
		\#types & 3368 & 7276 & 4102 & 4368 & 4778 \\
        increase & 1.1X & 3.9X & 4.1X & 2.4X & 2.9X \\
		\hline
	\end{tabular}
\end{table}
\smallskip
\noindent\textbf{Type triples:}\xspace
Table~\ref{tab:res-types} shows the quality of product-type triples measured by MTurk workers on 300 labels per domain. MoE shows the margin of error with a confidence level of 95\%. \method\ obtained an average precision of 87.7\% and increased the number of types in each domain by 2.9X on average.

\begin{table}[t]
	\centering
	\caption{\graph{} improves over input data on average by 7.6\% (percentage point) on precision, 16.4\% on recall, and 14.4\% on defect rate, with average MoE of 6.56\% on precision/recall. \label{tab:res-prototype} }
\vspace{-.1in}
	\resizebox{\columnwidth}{!}{
	\begin{tabular}{|c|cc|cc|cc|}
		\hline
                 & \multicolumn{2}{c|}{Attribute 1} & \multicolumn{2}{c|}{Attribute 2} & \multicolumn{2}{c|}{Attribute 3}        \\
                 		\hline
	\bf{Grocery}        	& Input     & \graph       & Input     & \graph & Input & \graph \\
	\hline

	Applicability   & \multicolumn{2}{c|}{38.51\%} &   \multicolumn{2}{c|}{7.53\%} &   \multicolumn{2}{c|}{10.00\%}     \\
	Precision       &  68.61\%           & 82.59\%           &     49.94\%        &    77.30\%   &   55.10\%      &  55.10\%   \\
	Recall          &     37.17\%       &    83.15\%        &      1.43\%      &   80.96\%    &  54.58\%     &    54.59\%   \\
        F-measure       &   48.22\%          &     82.87\%        &     2.78\%         &  79.09\%     &   54.84\%      &  54.85\%     \\
        Defect Rate     &   62.91\%          &     21.14\%        &     98.58\%        &    30.72\%   & 49.50\%        &   49.49\%    \\
	\hline
\bf{Health}     & Input     & \graph       & Input     & \graph & Input & \graph \\
\hline
Applicability   &   \multicolumn{2}{c|}{1.35\%}  & \multicolumn{2}{c|}{0.59\%}  &   \multicolumn{2}{c|}{57.92\%}    \\
Precision       &    70.54\%         &     84.00\%       &      59.11\%       &   70.00\%    &  78.00\%     &   61.75\%    \\
Recall          &      59.69\%       &     49.92\%    &     69.50\%        &  63.45\%     &  47.13\%       &   69.92\%   \\
F-measure       &   64.66\%          &     62.62\%        &     63.89\%         &  66.56\%     &     58.76\%    & 65.58\%      \\
Defect Rate     &     49.69\%        &  52.34\%           & 48.31\%            &     45.45\%  &   55.04\%      &  38.28\%     \\

\hline
\bf{Beauty}     & Input     & \graph       & Input     & \graph & Input & \graph \\
\hline

Applicability   &  \multicolumn{2}{c|}{0.04\%}   &  \multicolumn{2}{c|}{0.54\%} & \multicolumn{2}{c|}{4.82\%}     \\
Precision       & 18.83\%    &    48.00\%      &  71.98\%       & 76.00\%    & 62.00\%      &   61.68\%   \\
Recall          &     69.44\%       &    69.44\%      &     65.21\%       &  59.95\%     &  53.26\%       &  62.98\%     \\
F-measure       &   29.62\%          &     56.76\%        &     68.43\%         &  67.03\%     &   57.30\%      & 62.32\%      \\
Defect Rate     &    82.35\%         &   59.02\%          &  40.19\%           & 42.76\%      &  48.51\%       &  39.50\%     \\
\hline
\bf{Baby}       & Input     & \graph       & Input     & \graph & Input & \graph \\
	\hline
Applicability   &  \multicolumn{2}{c|}{0.0011\%}  &   \multicolumn{2}{c|}{0.09\%} & \multicolumn{2}{c|}{55.82\%}     \\
Precision       &    1.45\%     &     0.03\%      &    8.22\%         & 10.60\%     &  42.00\%     &  49.54\%    \\
Recall          &      9.79\%       &    9.79\%         &      3.83\%      &    50.92\%   &  44.13\%       &  56.39\%    \\
F-measure       &   2.53\%          &     0.06\%        &     5.23\%         &  17.55\%     &  43.04\%       &  52.74\%     \\
Defect Rate     &    98.72\%         &      99.97\%       &      97.30\%       &   89.63\%    &  60.81\%       &  50.46\%     \\

\hline
	\end{tabular}}
\end{table}

\smallskip
\noindent\textbf{Attribute triples:}\xspace
We start with choosing three text-valued attributes that are fairly popular to all 4 domains, and evaluated each (domain, attribute) pairs on 200 samples (Table~\ref{tab:res-prototype}). Note that even though they are popular among all text-valued attributes except {\em brand}, the applicability is still fairly low ($<$10\% most of the cases), showing the big diversity of each domain. We made three observations. First, we significantly increased the quality of the data (precision up by 7.6\%, recall up by 16.4\%, F-measure up by 14.1\%, and defect rate down by 14.4\%). Second, the quality of the generated data is often correlated with the precision of the raw data. For example, the precision of the data in the Baby domain is very low; as a result, the \graph\ data also have low precision. On the other hand, recall tends to have less effect; for example, Attribute 2 has a recall of 1.4\% for Grocery, but we are able to boost it to 81.0\% with reasonable precision (77.3\%). Third, there is still huge space for improvement: the defect rate is at best 21.1\%, and can be over 90\% in certain cases. We also note that production requires high accuracy, so we trade recall with precision in the production system.

\eat{
\begin{table}[htbp]
	\centering
	\caption{\method{} results in higher precision and coverage compared to the quality of the input data. Evaluation on several attributes specific for product type Coffee. \label{tab:res-polaris} }
        \vspace{-.1in}
	\resizebox{\columnwidth}{!}{
	\begin{tabular}{|l|c|c|c|c|}
		\hline
		{\bf Scope} & {\bf Input Prec} & {\bf Input Cov} & {\bf KG Prec} & {\bf KG Cov} \\
		\hline
		attribute-1 & 95.01\% & 34.23\%& 95.53\%& 44.07\%\\
		attribute-2 & 89.91\%& 46.99\%& 92.93\%& 62.06\%\\
		attribute-3 & 87.90\%& 0.3\%& 93.33\%& 21.46\%\\
		attribute-4 & 98.97\%& 7.72\%& 92.11\% & 81.30\%\\
		\hline
	\end{tabular}}
\end{table}
}

\eat{
\begin{table}[t]
	\centering
	\caption{\method{} obtained an average precision of 95.0\% and improved the recall by 4.3X for important categorical/binary attributes. \label{tab:res-polaris} }
        \vspace{-.1in}
	\resizebox{0.6\columnwidth}{!}{
	\begin{tabular}{|l|c|c|c|}
		\hline
		{\bf Scope} & {\bf Prec} & {\bf Recall} & {\bf Recall gain}\\
		\hline
		pair-1 & 91.05\% & 70.18\% & 7.3X \\
		pair-2 & 97.06\% & 19.70\% & 5.2X \\
		pair-3 & 97.12\% & 36.13\% & 1.3X \\
		pair-4 & 93.87\% & 37.72\% & 10.5X \\
                pair-5 & 95.88\% & 25.01\% & 1.2X \\
                pair-6 & 90.42\% & 87.46\% & 2.8X \\
                pair-7 & 97.97\% & 55.95\% & 1.4X \\
                pair-8 & 96.44\% & 87.49\% & 4.5X \\
		\hline
	\end{tabular}}
        \vspace{-.2in}
\end{table}
}

Next, we randomly chose 8 (type, attr) pairs for top-5 important attributes to report our results at a finer granularity (Table~\ref{tab:res-polaris} in Appendix~\ref{sec:appendix-example}). The attribute values are categorical (much smaller vocabulary) or binary, leading to higher extraction quality. \method\ obtained an average precision of 95.0\% and improved the recall by 4.3X. 

\begin{table}[t]
	\centering
	\caption{\method{} cleaned 1.77M incorrect values for two attributes with a precision of 90\%. \label{tab:res-cleaning} }
        \vspace{-.1in}
	\resizebox{\columnwidth}{!}{
	\begin{tabular}{|c|c|c|c|c|}
		\hline
		{\bf Grocery} & {\bf Recall@ 90\%} & {\bf Recall@ 80\%}  & {\bf \#Removed} & {\bf \%Removed} \\
		\hline
		Attribute 1 & 36.58\%& 58.20\%& 1,381,277& 55.06\%\\
		Attribute 2 & 9.92\% & 13.29\%& 320,960& 59.40\%\\
		\hline
		{\bf Health} & {\bf Recall@ 90\%} & {\bf Recall@ 80\%}  & {\bf \#Removed} & {\bf \%Removed} \\
		\hline
		Attribute 1 & 76.72\%& 85.93\%& 30,215& 32.63\%\\
		Attribute 2 & 3.18\%& 21.14\%& 14,110& 20.92\%\\
		\hline
		{\bf Beauty} & {\bf Recall@ 90\%} & {\bf Recall@ 80\%}  & {\bf \#Removed} & {\bf \%Removed} \\
		\hline
		Attribute 1 & 94.33\%& 97.16\%& 10,926& 62.31\%\\
		Attribute 2 & 46.05\%& 69.74\%& 7,651& 11.87\%\\
		\hline
		{\bf Baby} & {\bf Recall@ 90\%} & {\bf Recall@ 80\%}  & {\bf \#Removed} & {\bf \%Removed} \\
		\hline
		Attribute 1 & 87.24\%& 95.31\%& 2,673& 66.17\%\\
		Attribute 2 & 52.07\%& 59.24\%& 1,956& 73.04\%\\
		\hline
	\end{tabular}}
        \vspace{-.1in}
\end{table}

To highlight how data cleaning improves the data quality, we show in Table~\ref{tab:res-cleaning} the noisy values we have removed for the same two attributes as in Table~\ref{tab:res-prototype}. At a precision of 90\% (\ie, 9 out of 10 removed values are indeed incorrect), we achieve a recall of 73.7\% for Attribute 1 and 27.8\% for Attribute 2. In total we removed 1.3M values for these two attributes, accounting for 21\% of Catalog values and 64.6\% of \ourImpute.

\begin{table}
	\caption{Precision for triples representing relations between entities. Precision for synonym relations is reported on two representative attributes. \label{tab:res-rel} }
        \vspace{-.1in}
	\begin{tabular}{|c|c|c|c|}
		\hline
		& {\bf Attribute 1}  & {\bf Attribute 2} & {\bf Product Types} \\
		\hline
		Precision & 91.6\%  & 93.7\% & 88.1\% \\
                \#Pairs & 6,610 & 1,066    & 21,900 \\
		\hline
	\end{tabular}
        \vspace{-.2in}
\end{table}
\smallskip
\noindent\textbf{Relation triples:}\xspace
Finally, we show precision of relation triples in Table~\ref{tab:res-rel}, including hypernym relations between product types and synonym relations between attribute values. We observed very high precision for value synonyms ($>$90\%) and fairly high for hypernyms (88.1\%) when we consider attaching to any ancestor node (not necessarily to the leaf) as correct.

\section{Lessons We Learnt}
\label{sec:disc}
There are many lessons in building the Product KG, pointing to interesting research directions. 

{\bf Non-tree product hierarchies:} First, we may need to fundamentally reconsider the way we model taxonomy and classify products. Common practice is to structure product types into a tree structure for ``subtypes'', and classify each product into a single (ideally leaf) type. However, we miss multiple parents; for example, ``knife'' has subtypes ``chef's knife'', ``hunting knife'', which correspondingly is also a subtype of ``Cutlery \& Knife'' and ``Hunting kits''. Also, there is often no clear cut for product types: one product can be both fashion swimwear and two-piece swimwear. In general, we need to extend concepts to model broadly {\em subtype}, {\em synonym}, and {\em overlapping} relationships; for each product, we can simply extract the type from its title, and infer other types according to the relationship between product types.

{\bf Noisy data:} Second, the large volume of noises can deteriorate the performance of the imputation and cleaning models. This can be observed from our lower quality of knowledge in the Baby domain, caused by wrong product types and many inapplicable values in Catalog. We propose aggressive data cleaning before using the data for training, and training a multi-task end-to-end model that imputes missing values and identifies wrong or inapplicable values.

{\bf More than text:} Third, product profile is not the only source of product knowledge. A study on randomly sampled 22 (type, attribute) pairs shows that 71.3\% values can be derived from Amazon product profiles, an additional 3.8\%  can be derived from Amazon product images, and 24.9\% have to be sought from external sources such as manufacturer websites. This observation hints that a promising direction is to enhance \method\ with image processing capabilities and web extraction.

\section{Related Work}
\label{sec:related}

Industry KG systems typically rely on manually defined ontology and curate knowledge from Wikipedia and a number of structured sources (\eg, Freebase\cite{bollacker2008freebase}, Bing Satori~\cite{Gao2018Satori}, YAGO~\cite{fabian2007yago},
YAGO2~\cite{hoffart2013yago2}). Research systems conduct web extraction, but again observing pre-defined ontology and focus on named entities such as people, movies, companies (\eg, NELL~\cite{carlson2010toward}, Knowledge Vault~\cite{dong2014knowledge}, DeepDive~\cite{de2016deepdive}). By comparison, this work extracts product knowledge from product profiles, where the structure, sparsity and noise level are very different from webpages; many attribute values are free texts or numbers instead of named entities. Incorporating taxonomy knowledge into machine learning models and utilizing customer behavior signals for supervision are two themes employed throughout this work to improve model performance. 

The product knowledge graph described in~\cite{xu2020product} 
differs from our work as it focuses
on training product embeddings to represent
co-view/complement/substitute relationship defined therein, while this
work focuses on collecting factual knowledge about products (\eg, product types and attribute values).
Recent product property extraction systems~\cite{zheng2018opentag, xu2019scaling} apply tagging on product profiles, but consider a single product type.
Web extraction systems~\cite{rezk2019, qiu2015dexter} extract product knowledge from semi-structured websites, and the techniques are orthogonal to ours.

In addition to end-to-end systems, there have been solutions for individual components, including ontology
definition~\cite{fabian2007yago,bollacker2008freebase}, entity
identification~\cite{fabian2007yago}, relation
extraction~\cite{Lockard:2018:CDS:3231751.3242930}, hierarchical
embedding~\cite{nickel2017poincare},
linkage~\cite{papadakis2016comparative,gulhane2011web}, and knowledge
fusion~\cite{dong2014knowledge, dong2014fusion}. We apply these
techniques whenever appropriate, and improve them to address the
unique challenges for the product domain.

\section{Conclusions}
\label{sec:concl}
This paper describes our experience in building a broad knowledge graph for products of thousands of types. We applied a suite of ML methods to automate ontology construction, knowledge enrichment and cleaning for a large number of products with frequent changes. 
With these techniques we built a knowledge graph that significantly improves completeness, accuracy, and consistency of data comparing to Catalog. Our efforts also shed light on how we may further improve by going both broader and deeper in product graph construction.

\bibliographystyle{ACM-Reference-Format}
\bibliography{self-driving-pu-kdd}


\begin{thebibliography}{37}


\ifx \showCODEN    \undefined \def \showCODEN     #1{\unskip}     \fi
\ifx \showDOI      \undefined \def \showDOI       #1{#1}\fi
\ifx \showISBNx    \undefined \def \showISBNx     #1{\unskip}     \fi
\ifx \showISBNxiii \undefined \def \showISBNxiii  #1{\unskip}     \fi
\ifx \showISSN     \undefined \def \showISSN      #1{\unskip}     \fi
\ifx \showLCCN     \undefined \def \showLCCN      #1{\unskip}     \fi
\ifx \shownote     \undefined \def \shownote      #1{#1}          \fi
\ifx \showarticletitle \undefined \def \showarticletitle #1{#1}   \fi
\ifx \showURL      \undefined \def \showURL       {\relax}        \fi
\providecommand\bibfield[2]{#2}
\providecommand\bibinfo[2]{#2}
\providecommand\natexlab[1]{#1}
\providecommand\showeprint[2][]{arXiv:#2}

\bibitem[\protect\citeauthoryear{Abedjan, Chu, Deng, Fernandez, Ilyas, Ouzzani,
  Papotti, Stonebraker, and Tang}{Abedjan et~al\mbox{.}}{2016}]%
        {abedjan2016detecting}
\bibfield{author}{\bibinfo{person}{Ziawasch Abedjan}, \bibinfo{person}{Xu Chu},
  \bibinfo{person}{Dong Deng}, \bibinfo{person}{Raul~Castro Fernandez},
  \bibinfo{person}{Ihab~F Ilyas}, \bibinfo{person}{Mourad Ouzzani},
  \bibinfo{person}{Paolo Papotti}, \bibinfo{person}{Michael Stonebraker}, {and}
  \bibinfo{person}{Nan Tang}.} \bibinfo{year}{2016}\natexlab{}.
\newblock \showarticletitle{Detecting data errors: Where are we and what needs
  to be done?}
\newblock \bibinfo{journal}{\emph{Proceedings of the VLDB Endowment}}
  \bibinfo{volume}{9}, \bibinfo{number}{12} (\bibinfo{year}{2016}),
  \bibinfo{pages}{993--1004}.
\newblock


\bibitem[\protect\citeauthoryear{Bansal, Burkett, De~Melo, and Klein}{Bansal
  et~al\mbox{.}}{2014}]%
        {bansal2014structured}
\bibfield{author}{\bibinfo{person}{Mohit Bansal}, \bibinfo{person}{David
  Burkett}, \bibinfo{person}{Gerard De~Melo}, {and} \bibinfo{person}{Dan
  Klein}.} \bibinfo{year}{2014}\natexlab{}.
\newblock \showarticletitle{Structured Learning for Taxonomy Induction with
  Belief Propagation.}. In \bibinfo{booktitle}{\emph{ACL}}.
\newblock


\bibitem[\protect\citeauthoryear{Bojanowski, Grave, Joulin, and
  Mikolov}{Bojanowski et~al\mbox{.}}{2017}]%
        {bojanowski2017enriching}
\bibfield{author}{\bibinfo{person}{Piotr Bojanowski}, \bibinfo{person}{Edouard
  Grave}, \bibinfo{person}{Armand Joulin}, {and} \bibinfo{person}{Tomas
  Mikolov}.} \bibinfo{year}{2017}\natexlab{}.
\newblock \showarticletitle{Enriching word vectors with subword information}.
\newblock \bibinfo{journal}{\emph{TACL}}  \bibinfo{volume}{5}
  (\bibinfo{year}{2017}), \bibinfo{pages}{135--146}.
\newblock


\bibitem[\protect\citeauthoryear{Bollacker, Evans, Paritosh, Sturge, and
  Taylor}{Bollacker et~al\mbox{.}}{2008}]%
        {bollacker2008freebase}
\bibfield{author}{\bibinfo{person}{Kurt Bollacker}, \bibinfo{person}{Colin
  Evans}, \bibinfo{person}{Praveen Paritosh}, \bibinfo{person}{Tim Sturge},
  {and} \bibinfo{person}{Jamie Taylor}.} \bibinfo{year}{2008}\natexlab{}.
\newblock \showarticletitle{Freebase: a collaboratively created graph database
  for structuring human knowledge}. In \bibinfo{booktitle}{\emph{Sigmod}}. AcM,
  \bibinfo{pages}{1247--1250}.
\newblock


\bibitem[\protect\citeauthoryear{Bordea, Lefever, and Buitelaar}{Bordea
  et~al\mbox{.}}{2016}]%
        {bordea2016semeval}
\bibfield{author}{\bibinfo{person}{Georgeta Bordea}, \bibinfo{person}{Els
  Lefever}, {and} \bibinfo{person}{Paul Buitelaar}.}
  \bibinfo{year}{2016}\natexlab{}.
\newblock \showarticletitle{Semeval-2016 task 13: Taxonomy extraction
  evaluation (texeval-2)}. In \bibinfo{booktitle}{\emph{SemEval-2016}}.
  \bibinfo{pages}{1081--1091}.
\newblock


\bibitem[\protect\citeauthoryear{Carlson, Betteridge, Kisiel, Settles,
  Hruschka~Jr, and Mitchell}{Carlson et~al\mbox{.}}{2010}]%
        {carlson2010toward}
\bibfield{author}{\bibinfo{person}{Andrew Carlson}, \bibinfo{person}{Justin
  Betteridge}, \bibinfo{person}{Bryan Kisiel}, \bibinfo{person}{Burr Settles},
  \bibinfo{person}{Estevam~R Hruschka~Jr}, {and} \bibinfo{person}{Tom~M
  Mitchell}.} \bibinfo{year}{2010}\natexlab{}.
\newblock \showarticletitle{Toward an architecture for never-ending language
  learning.}. In \bibinfo{booktitle}{\emph{AAAI}}, Vol.~\bibinfo{volume}{5}.
  Atlanta, \bibinfo{pages}{3}.
\newblock


\bibitem[\protect\citeauthoryear{De~Sa, Ratner, R{\'e}, Shin, Wang, Wu, and
  Zhang}{De~Sa et~al\mbox{.}}{2016}]%
        {de2016deepdive}
\bibfield{author}{\bibinfo{person}{Christopher De~Sa}, \bibinfo{person}{Alex
  Ratner}, \bibinfo{person}{Christopher R{\'e}}, \bibinfo{person}{Jaeho Shin},
  \bibinfo{person}{Feiran Wang}, \bibinfo{person}{Sen Wu}, {and}
  \bibinfo{person}{Ce Zhang}.} \bibinfo{year}{2016}\natexlab{}.
\newblock \showarticletitle{Deepdive: Declarative knowledge base construction}.
\newblock \bibinfo{journal}{\emph{ACM SIGMOD Record}} \bibinfo{volume}{45},
  \bibinfo{number}{1} (\bibinfo{year}{2016}), \bibinfo{pages}{60--67}.
\newblock


\bibitem[\protect\citeauthoryear{Dong, Gabrilovich, Heitz, Horn, Lao, Murphy,
  Strohmann, Sun, and Zhang}{Dong et~al\mbox{.}}{2014a}]%
        {dong2014knowledge}
\bibfield{author}{\bibinfo{person}{Xin Dong}, \bibinfo{person}{Evgeniy
  Gabrilovich}, \bibinfo{person}{Geremy Heitz}, \bibinfo{person}{Wilko Horn},
  \bibinfo{person}{Ni Lao}, \bibinfo{person}{Kevin Murphy},
  \bibinfo{person}{Thomas Strohmann}, \bibinfo{person}{Shaohua Sun}, {and}
  \bibinfo{person}{Wei Zhang}.} \bibinfo{year}{2014}\natexlab{a}.
\newblock \showarticletitle{Knowledge vault: A web-scale approach to
  probabilistic knowledge fusion}. In \bibinfo{booktitle}{\emph{SigKDD}}. ACM,
  \bibinfo{pages}{601--610}.
\newblock


\bibitem[\protect\citeauthoryear{Dong, Gabrilovich, Heitz, Horn, Murphy, Sun,
  and Zhang}{Dong et~al\mbox{.}}{2014b}]%
        {dong2014fusion}
\bibfield{author}{\bibinfo{person}{Xin~Luna Dong}, \bibinfo{person}{Evgeniy
  Gabrilovich}, \bibinfo{person}{Geremy Heitz}, \bibinfo{person}{Wilko Horn},
  \bibinfo{person}{Kevin Murphy}, \bibinfo{person}{Shaohua Sun}, {and}
  \bibinfo{person}{Wei Zhang}.} \bibinfo{year}{2014}\natexlab{b}.
\newblock \showarticletitle{From Data Fusion to Knowledge Fusion}.
\newblock \bibinfo{journal}{\emph{PVLDB}} (\bibinfo{year}{2014}).
\newblock


\bibitem[\protect\citeauthoryear{Fabian, Gjergji, Gerhard,
  et~al\mbox{.}}{Fabian et~al\mbox{.}}{2007}]%
        {fabian2007yago}
\bibfield{author}{\bibinfo{person}{MS Fabian}, \bibinfo{person}{K Gjergji},
  \bibinfo{person}{Weikum Gerhard}, {et~al\mbox{.}}}
  \bibinfo{year}{2007}\natexlab{}.
\newblock \showarticletitle{Yago: A core of semantic knowledge unifying wordnet
  and wikipedia}. In \bibinfo{booktitle}{\emph{WWW}}.
  \bibinfo{pages}{697--706}.
\newblock


\bibitem[\protect\citeauthoryear{Furche, Gottlob, Grasso, Gunes, Guo,
  Kravchenko, Orsi, Schallhart, Sellers, and Wang}{Furche
  et~al\mbox{.}}{2012}]%
        {furche2012diadem}
\bibfield{author}{\bibinfo{person}{Tim Furche}, \bibinfo{person}{Georg
  Gottlob}, \bibinfo{person}{Giovanni Grasso}, \bibinfo{person}{Omer Gunes},
  \bibinfo{person}{Xiaoanan Guo}, \bibinfo{person}{Andrey Kravchenko},
  \bibinfo{person}{Giorgio Orsi}, \bibinfo{person}{Christian Schallhart},
  \bibinfo{person}{Andrew Sellers}, {and} \bibinfo{person}{Cheng Wang}.}
  \bibinfo{year}{2012}\natexlab{}.
\newblock \showarticletitle{DIADEM: domain-centric, intelligent, automated data
  extraction methodology}. In \bibinfo{booktitle}{\emph{WWW}}. ACM,
  \bibinfo{pages}{267--270}.
\newblock


\bibitem[\protect\citeauthoryear{Gao, Liang, Han, Yakout, and Mohamed}{Gao
  et~al\mbox{.}}{2018}]%
        {Gao2018Satori}
\bibfield{author}{\bibinfo{person}{Yuqing Gao}, \bibinfo{person}{Jisheng
  Liang}, \bibinfo{person}{Benjamin Han}, \bibinfo{person}{Mohamed Yakout},
  {and} \bibinfo{person}{Ahmed Mohamed}.} \bibinfo{year}{2018}\natexlab{}.
\newblock \showarticletitle{Building a large-scale, accurate and fresh
  knowledge graph}. In \bibinfo{booktitle}{\emph{SigKDD}}.
\newblock


\bibitem[\protect\citeauthoryear{Gulhane, Madaan, Mehta, Ramamirtham, Rastogi,
  Satpal, Sengamedu, Tengli, and Tiwari}{Gulhane et~al\mbox{.}}{2011}]%
        {gulhane2011web}
\bibfield{author}{\bibinfo{person}{Pankaj Gulhane}, \bibinfo{person}{Amit
  Madaan}, \bibinfo{person}{Rupesh Mehta}, \bibinfo{person}{Jeyashankher
  Ramamirtham}, \bibinfo{person}{Rajeev Rastogi}, \bibinfo{person}{Sandeep
  Satpal}, \bibinfo{person}{Srinivasan~H Sengamedu}, \bibinfo{person}{Ashwin
  Tengli}, {and} \bibinfo{person}{Charu Tiwari}.}
  \bibinfo{year}{2011}\natexlab{}.
\newblock \showarticletitle{Web-scale information extraction with Vertex}. In
  \bibinfo{booktitle}{\emph{ICDE}}. \bibinfo{pages}{1209--1220}.
\newblock


\bibitem[\protect\citeauthoryear{Hoffart, Suchanek, Berberich, and
  Weikum}{Hoffart et~al\mbox{.}}{2013}]%
        {hoffart2013yago2}
\bibfield{author}{\bibinfo{person}{Johannes Hoffart}, \bibinfo{person}{Fabian~M
  Suchanek}, \bibinfo{person}{Klaus Berberich}, {and} \bibinfo{person}{Gerhard
  Weikum}.} \bibinfo{year}{2013}\natexlab{}.
\newblock \showarticletitle{YAGO2: A spatially and temporally enhanced
  knowledge base from Wikipedia}.
\newblock \bibinfo{journal}{\emph{Artificial Intelligence}}
  \bibinfo{volume}{194} (\bibinfo{year}{2013}), \bibinfo{pages}{28--61}.
\newblock


\bibitem[\protect\citeauthoryear{Hopkinson, Gurdasani, Palfrey, and
  Mittal}{Hopkinson et~al\mbox{.}}{2018}]%
        {hopkinson2018demand}
\bibfield{author}{\bibinfo{person}{Andrew Hopkinson}, \bibinfo{person}{Amit
  Gurdasani}, \bibinfo{person}{Dave Palfrey}, {and} \bibinfo{person}{Arpit
  Mittal}.} \bibinfo{year}{2018}\natexlab{}.
\newblock \showarticletitle{Demand-Weighted Completeness Prediction for a
  Knowledge Base}. In \bibinfo{booktitle}{\emph{NAACL}}.
  \bibinfo{pages}{200--207}.
\newblock


\bibitem[\protect\citeauthoryear{Karamanolakis, Ma, and Dong}{Karamanolakis
  et~al\mbox{.}}{2020}]%
        {karamanolakis2020txtract}
\bibfield{author}{\bibinfo{person}{Giannis Karamanolakis}, \bibinfo{person}{Jun
  Ma}, {and} \bibinfo{person}{Xin~Luna Dong}.} \bibinfo{year}{2020}\natexlab{}.
\newblock \showarticletitle{TXtract: Taxonomy-Aware Knowledge Extraction for
  Thousands of Product Categories}. In \bibinfo{booktitle}{\emph{Proceedings of
  the 58th Annual Meeting of the Association for Computational Linguistics}}.
\newblock


\bibitem[\protect\citeauthoryear{Linden, Smith, and York}{Linden
  et~al\mbox{.}}{2003}]%
        {linden2003amazon}
\bibfield{author}{\bibinfo{person}{Greg Linden}, \bibinfo{person}{Brent Smith},
  {and} \bibinfo{person}{Jeremy York}.} \bibinfo{year}{2003}\natexlab{}.
\newblock \showarticletitle{Amazon. com recommendations: Item-to-item
  collaborative filtering}.
\newblock \bibinfo{journal}{\emph{IEEE Internet computing}}
  \bibinfo{volume}{7}, \bibinfo{number}{1} (\bibinfo{year}{2003}),
  \bibinfo{pages}{76--80}.
\newblock


\bibitem[\protect\citeauthoryear{Liu, Ting, and Zhou}{Liu
  et~al\mbox{.}}{2008}]%
        {liu2008isolation}
\bibfield{author}{\bibinfo{person}{Fei~Tony Liu}, \bibinfo{person}{Kai~Ming
  Ting}, {and} \bibinfo{person}{Zhi-Hua Zhou}.}
  \bibinfo{year}{2008}\natexlab{}.
\newblock \showarticletitle{Isolation forest}. In
  \bibinfo{booktitle}{\emph{2008 Eighth IEEE International Conference on Data
  Mining}}. IEEE, \bibinfo{pages}{413--422}.
\newblock


\bibitem[\protect\citeauthoryear{Liu, He, Chen, and Gao}{Liu
  et~al\mbox{.}}{2019}]%
        {liu2019mt-dnn}
\bibfield{author}{\bibinfo{person}{Xiaodong Liu}, \bibinfo{person}{Pengcheng
  He}, \bibinfo{person}{Weizhu Chen}, {and} \bibinfo{person}{Jianfeng Gao}.}
  \bibinfo{year}{2019}\natexlab{}.
\newblock \showarticletitle{Multi-Task Deep Neural Networks for Natural
  Language Understanding}. In \bibinfo{booktitle}{\emph{ACL}}.
  \bibinfo{pages}{4487--4496}.
\newblock


\bibitem[\protect\citeauthoryear{Lockard, Dong, Einolghozati, and
  Shiralkar}{Lockard et~al\mbox{.}}{2018}]%
        {Lockard:2018:CDS:3231751.3242930}
\bibfield{author}{\bibinfo{person}{Colin Lockard}, \bibinfo{person}{Xin~Luna
  Dong}, \bibinfo{person}{Arash Einolghozati}, {and} \bibinfo{person}{Prashant
  Shiralkar}.} \bibinfo{year}{2018}\natexlab{}.
\newblock \showarticletitle{CERES: Distantly Supervised Relation Extraction
  from the Semi-structured Web}.
\newblock \bibinfo{journal}{\emph{PVLDB}} (\bibinfo{year}{2018}),
  \bibinfo{pages}{1084--1096}.
\newblock


\bibitem[\protect\citeauthoryear{Mao, Ren, Shen, Gu, and Han}{Mao
  et~al\mbox{.}}{2018}]%
        {mao2018end}
\bibfield{author}{\bibinfo{person}{Yuning Mao}, \bibinfo{person}{Xiang Ren},
  \bibinfo{person}{Jiaming Shen}, \bibinfo{person}{Xiaotao Gu}, {and}
  \bibinfo{person}{Jiawei Han}.} \bibinfo{year}{2018}\natexlab{}.
\newblock \showarticletitle{End-to-End Reinforcement Learning for Automatic
  Taxonomy Induction}. In \bibinfo{booktitle}{\emph{ACL}}.
  \bibinfo{pages}{2462--2472}.
\newblock


\bibitem[\protect\citeauthoryear{Mao, Zhao, Kan, Zhang, Dong, Faloutsos, and
  Han}{Mao et~al\mbox{.}}{2020}]%
        {mao2020octet}
\bibfield{author}{\bibinfo{person}{Yuning Mao}, \bibinfo{person}{Tong Zhao},
  \bibinfo{person}{Andrey Kan}, \bibinfo{person}{Chenwei Zhang},
  \bibinfo{person}{Xin~Luna Dong}, \bibinfo{person}{Christos Faloutsos}, {and}
  \bibinfo{person}{Jiawei Han}.} \bibinfo{year}{2020}\natexlab{}.
\newblock \showarticletitle{OCTET: Online Catalog Taxonomy Enrichment with
  Self-Supervision}. In \bibinfo{booktitle}{\emph{SigKDD}}.
\newblock


\bibitem[\protect\citeauthoryear{Nickel and Kiela}{Nickel and Kiela}{2017}]%
        {nickel2017poincare}
\bibfield{author}{\bibinfo{person}{Maximillian Nickel} {and}
  \bibinfo{person}{Douwe Kiela}.} \bibinfo{year}{2017}\natexlab{}.
\newblock \showarticletitle{Poincar{\'e} embeddings for learning hierarchical
  representations}. In \bibinfo{booktitle}{\emph{NIPS}}.
  \bibinfo{pages}{6338--6347}.
\newblock


\bibitem[\protect\citeauthoryear{Papadakis, Svirsky, Gal, and
  Palpanas}{Papadakis et~al\mbox{.}}{2016}]%
        {papadakis2016comparative}
\bibfield{author}{\bibinfo{person}{George Papadakis}, \bibinfo{person}{Jonathan
  Svirsky}, \bibinfo{person}{Avigdor Gal}, {and} \bibinfo{person}{Themis
  Palpanas}.} \bibinfo{year}{2016}\natexlab{}.
\newblock \showarticletitle{Comparative analysis of approximate blocking
  techniques for entity resolution}.
\newblock \bibinfo{journal}{\emph{Proceedings of the VLDB Endowment}}
  \bibinfo{volume}{9}, \bibinfo{number}{9} (\bibinfo{year}{2016}),
  \bibinfo{pages}{684--695}.
\newblock


\bibitem[\protect\citeauthoryear{Pennington, Socher, and Manning}{Pennington
  et~al\mbox{.}}{2014}]%
        {pennington2014glove}
\bibfield{author}{\bibinfo{person}{Jeffrey Pennington},
  \bibinfo{person}{Richard Socher}, {and} \bibinfo{person}{Christopher~D.
  Manning}.} \bibinfo{year}{2014}\natexlab{}.
\newblock \showarticletitle{GloVe: Global Vectors for Word Representation}. In
  \bibinfo{booktitle}{\emph{EMNLP}}. \bibinfo{pages}{1532--1543}.
\newblock


\bibitem[\protect\citeauthoryear{Qiu, Barbosa, Dong, Shen, and Srivastava}{Qiu
  et~al\mbox{.}}{2015}]%
        {qiu2015dexter}
\bibfield{author}{\bibinfo{person}{Disheng Qiu}, \bibinfo{person}{Luciano
  Barbosa}, \bibinfo{person}{Xin~Luna Dong}, \bibinfo{person}{Yanyan Shen},
  {and} \bibinfo{person}{Divesh Srivastava}.} \bibinfo{year}{2015}\natexlab{}.
\newblock \showarticletitle{Dexter: large-scale discovery and extraction of
  product specifications on the web}.
\newblock \bibinfo{journal}{\emph{Proceedings of the VLDB Endowment}}
  \bibinfo{volume}{8}, \bibinfo{number}{13} (\bibinfo{year}{2015}),
  \bibinfo{pages}{2194--2205}.
\newblock


\bibitem[\protect\citeauthoryear{Razniewski, Balaraman, and Nutt}{Razniewski
  et~al\mbox{.}}{2017}]%
        {razniewski2017doctoral}
\bibfield{author}{\bibinfo{person}{Simon Razniewski}, \bibinfo{person}{Vevake
  Balaraman}, {and} \bibinfo{person}{Werner Nutt}.}
  \bibinfo{year}{2017}\natexlab{}.
\newblock \showarticletitle{Doctoral advisor or medical condition: Towards
  entity-specific rankings of knowledge base properties}. In
  \bibinfo{booktitle}{\emph{International Conference on Advanced Data Mining
  and Applications}}.
\newblock


\bibitem[\protect\citeauthoryear{Rezk, Alemany, Nio, and Zhang}{Rezk
  et~al\mbox{.}}{2019}]%
        {rezk2019}
\bibfield{author}{\bibinfo{person}{Martin Rezk}, \bibinfo{person}{Laura~Alonso
  Alemany}, \bibinfo{person}{Lasguido Nio}, {and} \bibinfo{person}{Ted Zhang}.}
  \bibinfo{year}{2019}\natexlab{}.
\newblock \showarticletitle{Accurate product attribute extraction on the
  field}. In \bibinfo{booktitle}{\emph{ICDE}}. \bibinfo{pages}{1862–1873}.
\newblock


\bibitem[\protect\citeauthoryear{Schlichtkrull and Alonso}{Schlichtkrull and
  Alonso}{2016}]%
        {schlichtkrull2016msejrku}
\bibfield{author}{\bibinfo{person}{Michael Schlichtkrull} {and}
  \bibinfo{person}{H{\'e}ctor~Mart{\'\i}nez Alonso}.}
  \bibinfo{year}{2016}\natexlab{}.
\newblock \showarticletitle{Msejrku at semeval-2016 task 14: Taxonomy
  enrichment by evidence ranking}. In \bibinfo{booktitle}{\emph{SemEval}}.
\newblock


\bibitem[\protect\citeauthoryear{Shang, Liu, Jiang, Ren, Voss, and Han}{Shang
  et~al\mbox{.}}{2018}]%
        {shang2018automated}
\bibfield{author}{\bibinfo{person}{Jingbo Shang}, \bibinfo{person}{Jialu Liu},
  \bibinfo{person}{Meng Jiang}, \bibinfo{person}{Xiang Ren},
  \bibinfo{person}{Clare~R Voss}, {and} \bibinfo{person}{Jiawei Han}.}
  \bibinfo{year}{2018}\natexlab{}.
\newblock \showarticletitle{Automated phrase mining from massive text corpora}.
\newblock \bibinfo{journal}{\emph{IEEE Transactions on Knowledge and Data
  Engineering}} \bibinfo{volume}{30}, \bibinfo{number}{10}
  (\bibinfo{year}{2018}), \bibinfo{pages}{1825--1837}.
\newblock


\bibitem[\protect\citeauthoryear{Sun, Yang, Zhang, Chen, Wei, Meng, and Hu}{Sun
  et~al\mbox{.}}{2018}]%
        {sun2018important}
\bibfield{author}{\bibinfo{person}{Shengjie Sun}, \bibinfo{person}{Dong Yang},
  \bibinfo{person}{Hongchun Zhang}, \bibinfo{person}{Yanxu Chen},
  \bibinfo{person}{Chao Wei}, \bibinfo{person}{Xiaonan Meng}, {and}
  \bibinfo{person}{Yi Hu}.} \bibinfo{year}{2018}\natexlab{}.
\newblock \showarticletitle{Important Attribute Identification in Knowledge
  Graph}.
\newblock \bibinfo{journal}{\emph{arXiv preprint arXiv:1810.05320}}
  (\bibinfo{year}{2018}).
\newblock


\bibitem[\protect\citeauthoryear{Vaswani, Shazeer, Parmar, Uszkoreit, Jones,
  Gomez, Kaiser, and Polosukhin}{Vaswani et~al\mbox{.}}{2017}]%
        {vaswani2017attention}
\bibfield{author}{\bibinfo{person}{Ashish Vaswani}, \bibinfo{person}{Noam
  Shazeer}, \bibinfo{person}{Niki Parmar}, \bibinfo{person}{Jakob Uszkoreit},
  \bibinfo{person}{Llion Jones}, \bibinfo{person}{Aidan~N Gomez},
  \bibinfo{person}{{\L}ukasz Kaiser}, {and} \bibinfo{person}{Illia
  Polosukhin}.} \bibinfo{year}{2017}\natexlab{}.
\newblock \showarticletitle{Attention is all you need}. In
  \bibinfo{booktitle}{\emph{Advances in neural information processing
  systems}}. \bibinfo{pages}{5998--6008}.
\newblock


\bibitem[\protect\citeauthoryear{Wang, Kang, Chang, and Han}{Wang
  et~al\mbox{.}}{2014}]%
        {wang2014hierarchical}
\bibfield{author}{\bibinfo{person}{Jingjing Wang}, \bibinfo{person}{Changsung
  Kang}, \bibinfo{person}{Yi Chang}, {and} \bibinfo{person}{Jiawei Han}.}
  \bibinfo{year}{2014}\natexlab{}.
\newblock \showarticletitle{A hierarchical dirichlet model for taxonomy
  expansion for search engines}. In \bibinfo{booktitle}{\emph{WWW}}.
\newblock


\bibitem[\protect\citeauthoryear{Xu, Ruan, Korpeoglu, Kumar, and Achan}{Xu
  et~al\mbox{.}}{2020}]%
        {xu2020product}
\bibfield{author}{\bibinfo{person}{Da Xu}, \bibinfo{person}{Chuanwei Ruan},
  \bibinfo{person}{Evren Korpeoglu}, \bibinfo{person}{Sushant Kumar}, {and}
  \bibinfo{person}{Kannan Achan}.} \bibinfo{year}{2020}\natexlab{}.
\newblock \showarticletitle{Product Knowledge Graph Embedding for E-commerce}.
  In \bibinfo{booktitle}{\emph{Proceedings of the 13th International Conference
  on Web Search and Data Mining}}. \bibinfo{pages}{672--680}.
\newblock


\bibitem[\protect\citeauthoryear{Xu, Wang, Mao, Jiang, and Lan}{Xu
  et~al\mbox{.}}{2019}]%
        {xu2019scaling}
\bibfield{author}{\bibinfo{person}{Huimin Xu}, \bibinfo{person}{Wenting Wang},
  \bibinfo{person}{Xinnian Mao}, \bibinfo{person}{Xinyu Jiang}, {and}
  \bibinfo{person}{Man Lan}.} \bibinfo{year}{2019}\natexlab{}.
\newblock \showarticletitle{Scaling up Open Tagging from Tens to Thousands:
  Comprehension Empowered Attribute Value Extraction from Product Title}. In
  \bibinfo{booktitle}{\emph{ACL}}. \bibinfo{pages}{5214--5223}.
\newblock


\bibitem[\protect\citeauthoryear{Zhang, Mukherjee, Lockard, Dong, and
  McCallum}{Zhang et~al\mbox{.}}{2019}]%
        {zhang2019openki}
\bibfield{author}{\bibinfo{person}{Dongxu Zhang}, \bibinfo{person}{Subhabrata
  Mukherjee}, \bibinfo{person}{Colin Lockard}, \bibinfo{person}{Xin~Luna Dong},
  {and} \bibinfo{person}{Andrew McCallum}.} \bibinfo{year}{2019}\natexlab{}.
\newblock \showarticletitle{OpenKI: Integrating Open Information Extraction and
  Knowledge Bases with Relation Inference}.
\newblock \bibinfo{journal}{\emph{arXiv preprint arXiv:1904.12606}}
  (\bibinfo{year}{2019}).
\newblock


\bibitem[\protect\citeauthoryear{Zheng, Mukherjee, Dong, and Li}{Zheng
  et~al\mbox{.}}{2018}]%
        {zheng2018opentag}
\bibfield{author}{\bibinfo{person}{Guineng Zheng}, \bibinfo{person}{Subhabrata
  Mukherjee}, \bibinfo{person}{Xin~Luna Dong}, {and} \bibinfo{person}{Feifei
  Li}.} \bibinfo{year}{2018}\natexlab{}.
\newblock \showarticletitle{OpenTag: Open Attribute Value Extraction from
  Product Profiles}. In \bibinfo{booktitle}{\emph{SigKDD}}.
\newblock


\end{thebibliography}

\appendix
\newpage

\section{Examples}
\label{sec:appendix-example}
\begin{figure}
	\subfloat[]{%
		\includegraphics[clip,width=1\columnwidth]{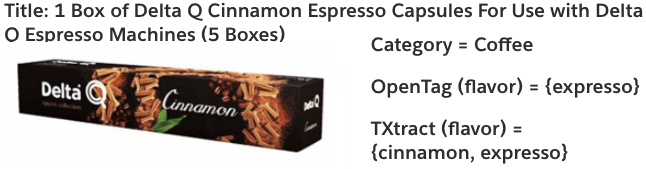}%
	}
	
	
	\subfloat[]{%
		\includegraphics[clip,width=1\columnwidth]{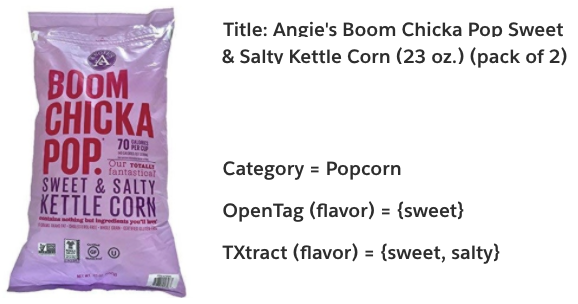}%
	}
	
	\subfloat[]{%
		\includegraphics[clip,width=1\columnwidth]{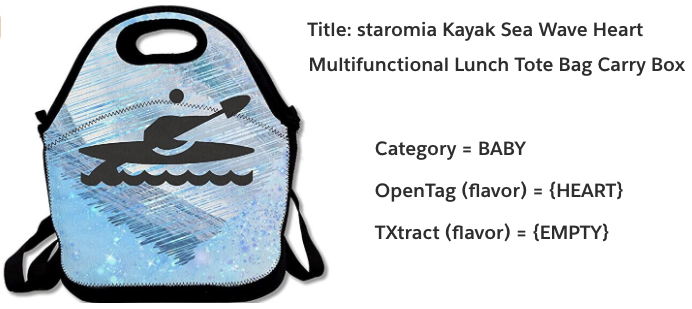}%
	}
	\caption{Examples of extracted attribute values from OpenTag and TXtract.\label{fig:eg:txtract}}
\end{figure}

Here we provide example outputs produced by each of the five components (Figure~\ref{fig:arch}). Taxonomy enrichment and relation discovery results are shown in Tables~\ref{tab:eg:types},~\ref{tab:eg:hypernyms} and~\ref{tab:eg:attributes}. Next, data imputation, cleaning and synonym finding results are shown in Figure~\ref{fig:eg:txtract} and Tables~\ref{tab:eg:cleaning} and~\ref{tab:eg:synonyms}. Finally, we also show additional evaluation results for the entire pipeline on a sample of type-attribute pairs in Table~\ref{tab:res-polaris} (see evaluation details in Section~\ref{sec:exp}).

\begin{table}[h]
	\caption{Examples of type extraction results.\label{tab:eg:types}}
	\begin{tabular}{|lp{2in}l|}
		\hline
		{\bf Source} & {\bf Text}                                                                                                          & {\bf Product Type}   \\ \hline
		Product      & {\em 4 Country Pasta Homemade Style Egg Pasta - 16-oz bag}                                                          & {\em Egg Pasta}      \\
		Product      & {\em Hamburger Helper Lasagna Pasta, Four Cheese, 10.3 Ounce (Pack of 6)}                                           & {\em Lasagna Pasta}  \\
		Product      & {\em COFFEE MATE The Original Powder Coffee Creamer 35.3 Oz. Canister Non-dairy, Lactose Free, Gluten Free Creamer} & {\em Coffee Creamer} \\
		Query        & {\em mccormick paprika 8.5 ounce}                                                                                   & {\em paprika}        \\
		Query        & {\em flax seeds raw}                                                                                                & {\em flax seeds}     \\ \hline
	\end{tabular}
\end{table}

\begin{table}[h]
	\caption{Examples of detected product type hypernyms.\label{tab:eg:hypernyms}}
	\begin{tabular}{|cc|}
		\hline
		{\bf Child Type}           & {\bf Parent Type}              \\
		\hline
		{\em Coconut flour}        & {\em Baking flours \& meals} \\
		{\em Tilapia}              & {\em Fresh fish}            \\
		{\em Fresh cut carnations} & {\em Fresh cut flowers}     \\
		{\em Bock beers}           & {\em Lager \& pilsner beers} \\
		{\em Pinto beans}          & {\em Dried beans}           \\
		\hline
	\end{tabular}
\end{table}

\begin{table}[h]
	\caption{Attributes identified as most important for two example types.\label{tab:eg:attributes}}
	\begin{tabular}{|c|c|}
		\hline
		{\bf Cereals} & {\bf Shampoo} \\
		\hline
		{\em brand} & {\em brand} \\
		{\em ingredients} & {\em hair type} \\
		{\em flavor} & {\em number of items} \\
		{\em number of items} & {\em ingredients} \\
		{\em energy content} & {\em liquid volume} \\
		\hline
	\end{tabular}
\end{table}

\begin{table}[htb]
	\centering
	\caption{\small Example errors found by the cleaning model.\label{tab:eg:cleaning}}
	\small
	\begin{tabular}{|p{0.55\linewidth}|p{0.11\linewidth}|p{0.19\linewidth}|}
		\hline
		{\bf Product profile} & {\bf Attrib.} & {\bf \thead{Attribute\\ value}} \\
		\hline
		Love of Candy Bulk Candy - Pink Mint Chocolate Lentils - 6lb Bag& Flavor & Pink \\
		\hline
		Scott's Cakes Dark Chocolate Fruit \& Nut Cream Filling Candies with Burgandy Foils in a 1 Pound Snowflake Box & Flavor & snowflake box \\
		\hline
		Lucky Baby - Baby Blanket Envelope Swaddle Winter Wrap Coral Fleece Newborn Blanket Sleeper Infant Stroller Wrap Toddlers Baby Sleeping Bag (color 1)& Flavor & Color 1 \\
		\hline
		ASUTRA Himalayan Sea Salt Body Scrub Exfoliator + Body Brush (Vitamin C), 12 oz | Ultra Hydrating, Gentle, Moisturizing | All Natural \& Organic Jojoba, Sweet Almond, Argan Oils & Scent & vitamin c body scrub - 12oz \& body brush \\
		\hline
		Folgers Simply Smooth Ground Coffee, 2 Count (Medium Roast), 31.1 Ounce& Scent & 2Packages (Breakfast Blend, 31.1 oz)\\
		
		\hline
	\end{tabular}
\end{table}

\begin{table}
	\caption{Examples of discovered flavor and scent synonym pairs.\label{tab:eg:synonyms}}
	\begin{tabular}{|cc|}
		\hline
		\multicolumn{2}{|c|}{flavor synonyms}  \\
		\hline
		herb and garlic & herb \& garlic \\
		macadamia nut & macadamia  \\
		roasted oolong tea & roasted oolong    \\
		decaffeinated honey lemon  & decaf honey lemon \\
		zero carb vanilla  & zero cal vanilla \\
		\hline
		\multicolumn{2}{|c|}{scent synonyms}  \\
		\hline
		basil (sweet) & sweet basil \\
		rose flower & rose \\
		aloe lubricant & aloe lube\\
		unscented & uncented\\
		moonlight path & moonlit path\\
		\hline
	\end{tabular}
\end{table}

\begin{table}
	\centering
	\caption{\method{} obtained an average precision of 95.0\% and improved the recall by 4.3X for important categorical/binary attributes.\label{tab:res-polaris}}
	\vspace{-.1in}
	\resizebox{0.6\columnwidth}{!}{
		\begin{tabular}{|l|c|c|c|}
			\hline
			{\bf Scope} & {\bf Prec} & {\bf Recall} & {\bf Recall gain}\\
			\hline
			pair-1 & 91.05\% & 70.18\% & 7.3X \\
			pair-2 & 97.06\% & 19.70\% & 5.2X \\
			pair-3 & 97.12\% & 36.13\% & 1.3X \\
			pair-4 & 93.87\% & 37.72\% & 10.5X \\
			pair-5 & 95.88\% & 25.01\% & 1.2X \\
			pair-6 & 90.42\% & 87.46\% & 2.8X \\
			pair-7 & 97.97\% & 55.95\% & 1.4X \\
			pair-8 & 96.44\% & 87.49\% & 4.5X \\
			\hline
	\end{tabular}}
	\vspace{-.2in}
\end{table}

\section{Attribute Applicability and Importance}
\label{sec:appendix-attribute}

Recall that for each (product type, attribute) pair we need to identify whether the attribute applies and how important the attribute is (e.g., whether {\em color} applies to {\em Shoes}, and how important is {\em color} for {\em Shoes}). To this end, we independently train a Random Forest classifier to predict applicability and a Random Forest Regressor to predict importance scores (for applicable attributes). In both cases, we consider each (product type, attribute) pair as an instance, and we label a sample of such pairs with either applicability or importance labels (Section~\ref{sec:attributes}). Sample prediction results for attribute importance are shown in Table~\ref{tab:eg:attributes}

In both models, we use the same set of features that characterize how relevant the attribute is for the given product type. The first feature is coverage, which is the proportion of products that have a non-missing attribute value. Next, a range of features are based on frequency of attribute mentions in different text sources. Consider a text source $s$ (e.g., product descriptions, reviews, etc.), and suppose that all products of the required type are indexed from $1$ to $n$. Note that a particular product $i$, can have several pieces of text of type $s$ (e.g., a product might have several reviews), and let $l(s,i)$ denote the number of such pieces. A feature based on signal $s$ is then defined as
$x(s) = \frac{1}{n}\sum_{i=1}^{n}\left(\frac{1}{l(s,i)}\sum_{j=1}^{l(s,i)}M(s,i,j)\right)$
Here $M(s,i,j)=1$ if the $j$-th piece of text of signal $s$ associated with product $i$ mentions the attribute (e.g., whether the $j$-th review of the $i$-th shoes mentions {\em color}), otherwise $M(s,i,j)=0$.

We consider two implementations of $M(s,i,j)$, and accordingly, for each $s$ we compute two features. First, $M(s,i,j)=1$ if text piece $j$ contains attribute value of product $i$ (e.g., whether the review for product $i$ contains {\em color} of this product). Second, $M(s,i,j)=1$ if text piece $j$ contains any common attribute value for this product type (\ie, whether the review for product $i$ contains any frequent color values among {\em Shoes}). We consider a value to be common if it is among the top 30 most frequent values within the given type. We consider several text signals (e.g., product titles, reviews, search queries, etc.) and compute $30$ features as described above. Finally, for each feature we also consider an alternative where products are weighted by popularity, and thus in total we have $60$ features.

\section{Taxonomy-Aware Semantic Cleaning}
\label{sec:appendix-cleaning}
\label{appendix:cleaning}
\newcommand{\bD}{\boldsymbol{D}}
\newcommand{\bV}{\boldsymbol{V}}
\newcommand{\bT}{\boldsymbol{T}}
\newcommand{\bS}{\boldsymbol{S}}
\newcommand{\be}{\boldsymbol{e}}

The cleaning model detects whether or not a triple $(\operatorname{PID}, A, V)$ is correct  (\ie, whether $V$ is the correct value of attribute $A$ for product $\operatorname{PID}$) by attending to its taxonomy node and semantic signals in product profile. Let $D=[d_1,\ldots,d_{n_D}]$, $T=[t_1,\ldots,t_{n_T}]$  and $V=[v_1,\ldots,v_{n_V}]$ be the token sequences of the product description, product taxonomy, and target attribute value, respectively. We construct the input sequence $S$ by concatenating $D$, $T$ and $V$ and inserting special tokens "$[CLS]$" and "$[SEP]$" as follows.
\begin{figure}[t]
  \centering
    \includegraphics[width=0.45\textwidth]{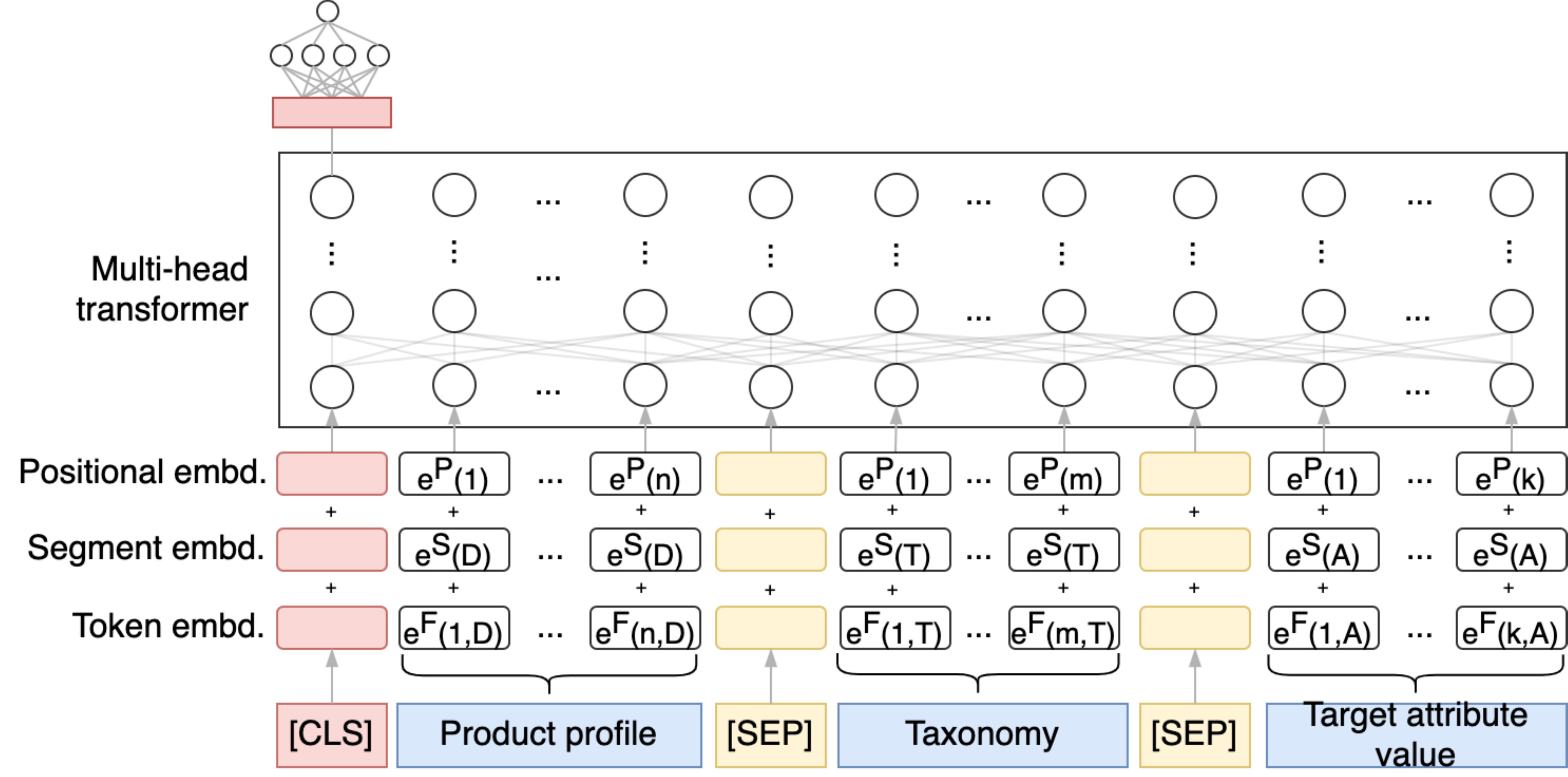}
     \caption{Cleaning model architecture.}
    \label{fig:cleaning}
\end{figure} 

\begin{equation}
\label{eq:cleaning-raw}
\bS = concat([CLS], D, [SEP],T, [SEP], V) := [s_1, \ldots, s_{n_S}]
\end{equation}
where $n_S = n_D + n_T + n_V + 3$. We then map each $s_i\in S$ to an embedding vector $\boldsymbol{e}_i\in \mathbb{R}^d$ as the summation of three embedding vectors of the same dimension $d$:

\begin{equation}
\label{eq:cleaning-input-embedding}
\be_i = \be^{\text{FastText}}_i + \be^{\text{Segment}}_i + \be^{\text{Position}}_i, i=1,\ldots,n_S
\end{equation}
where $\be^{\text{FastText}}_i$ is the pretrained FastText embedding~\cite{vaswani2017attention} of $s_i$, $\boldsymbol{e}^{\text{Segment}}_i$ is a segment embedding vector defined as:  
\begin{equation}
   \boldsymbol{e}^{\text{Segment}}_i =
   \begin{cases}
     \be^{\text{D}}, & \text{if}\ s_i\in D \\
     \be^{\text{T}}, & \text{if}\ s_i\in T \\
     \be^{\text{V}}, & \text{if}\ s_i\in V,
   \end{cases}
\end{equation}
and $\be^{\text{Position}}_i$ is the position embedding vector of the location of $s_i$ in the sequence (i.e. $i$), for which we adopt the same constructions used in ~\cite{vaswani2017attention}. Here $\be^{\text{FastText}}_i$'s and $\be^{\text{Position}}_i$'s are frozen (not trainable), and $\be^{\text{D}}$, $\be^{\text{T}}$, $\be^{\text{V}}$ are randomly initialized and jointly trained with other model parameters. 

The embedding sequence $[\be_i]_1^{n_S}$  is propagated through a multi-layer transformer model where number of layers, number of heads and hidden dimension are hyperparameters. The final embedding vector of the special token [CLS], denoted by $\boldsymbol{e}^{Out}$, captures the distilled representations of all three input sequences. It is passed through a dense layer followed by a sigmoid node to produce a single score between $0$ and $1$, indicating the likelihood of the input triple $(\operatorname{PID}, A, V)$ being correct. See Figure \ref{fig:cleaning} for an illustration of the model architecture. 

In Table \ref{tab:eg:cleaning} we give examples of attribute value errors detected by the cleaning model.

\end{document}